\documentclass[12pt]{article}
\usepackage{psfrag,epsf,color,subfigure}
\usepackage{enumerate}
\usepackage{natbib}
\usepackage[unicode=true,pdfusetitle,
   bookmarks=true, 
   bookmarksnumbered=false, 
   bookmarksopen=false,
   breaklinks=false, 
   backref=false, 
   colorlinks=false]{hyperref}
 \usepackage{mathtools}
\usepackage{amsthm,amsmath,amssymb,bbm,bm}
\usepackage{hyperref}
\usepackage{amssymb}
\usepackage{multirow}
\usepackage{array}
\usepackage{bm}
\usepackage{booktabs}
\usepackage{siunitx} 
\usepackage{float}
\usepackage{multirow}
\usepackage[pdftex]{graphicx}
\usepackage{algorithm}
\usepackage{algorithmic}
\usepackage{placeins}
\usepackage{mathtools}
\usepackage{romannum}
\usepackage{mathrsfs}
\usepackage{dsfont}
\usepackage{relsize}
\usepackage{rotating}
\usepackage{enumitem}
\usepackage{setspace}
\usepackage{minitoc}
\usepackage{etoc}
\usepackage{graphicx}
\usepackage{subcaption}
\usepackage{listings}
\usepackage[font=small,labelfont=bf]{caption}
\setcitestyle{citesep={;}}
\usepackage[left=0.7in, right=0.7in, top=0.9in, bottom=1.1in]{geometry}
\usepackage{hyperref}

\usepackage{changes}
\usepackage[utf8]{inputenc} 
\usepackage[T1]{fontenc}    
\usepackage{hyperref}       
\usepackage{url}            
\usepackage{booktabs}       
\usepackage{amsfonts}       
\usepackage{nicefrac}       
\usepackage{microtype}      
\usepackage{xcolor}         

\usepackage{psfrag,epsf,color}
\usepackage{amssymb}
\usepackage{epsfig}
\usepackage{subfigure} 

\usepackage{mathtools}

\DeclarePairedDelimiter\floor{\lfloor}{\rfloor}

\DeclareMathOperator*{\argmax}{arg\,max}

\usepackage{authblk}
\usepackage{dsfont}

\newtheorem{thm}{Theorem}

\counterwithin{thm}{section}

\counterwithin{example}{section}
\counterwithin{rmk}{section}
\counterwithin{proposition}{section}

\counterwithin{corollary}{section}
\newtheorem{assumption}{Assumption}[thm]
\counterwithin{assumption}{section}
\newtheorem{definition}{Definition}[thm]
\counterwithin{definition}{section}

\usepackage[page,header]{appendix}
\usepackage{titletoc}
\usepackage{lipsum}

\usepackage{caption} 
\def\T{{ \mathrm{\scriptscriptstyle T} }}

\makeatletter
\newcommand{\vast}{\bBigg@{4}}
\newcommand{\Vast}{\bBigg@{5}}
\makeatother

    \makeatletter
\def\@fnsymbol#1{\ensuremath{\ifcase#1\or *\or \mathparagraph\or
   \dagger\or \mathparagraph\or \|\or **\or \dagger\dagger
   \or \ddagger\ddagger \else\@ctrerr\fi}}
    \makeatother

\def\##1\#{\begin{align}#1\end{align}}
\def\$#1\${\begin{align*}#1\end{align*}}

\def\spacingset#1{\renewcommand{\baselinestretch}%
{#1}\small\normalsize} \spacingset{1}

\begin{document}

\title{\bf A Model-Agnostic Graph Neural Network for Integrating Local and Global Information}
\author[1]{Wenzhuo Zhou}
\author[1]{Annie Qu}
\author[2]{Keiland W. Cooper}
\author[2]{Norbert Fortin}
\author[1]{\\Babak Shahbaba}
\affil[1]{Department of Statistics, University of California Irvine}
\affil[2]{Department of Neurobiology and Behavior, University of California Irvine}
\date{}

\maketitle

\setcounter{page}{1}
\pagenumbering{arabic}

\abstract{
Graph Neural Networks (GNNs) have achieved promising performance in a variety of graph-focused tasks.  
Despite their success, however, existing GNNs suffer from two significant limitations: a lack of interpretability in their results due to their black-box nature, and an inability to learn representations of varying orders. To tackle these issues, we propose a novel \textbf{M}odel-\textbf{a}gnostic \textbf{G}raph Neural \textbf{Net}work (MaGNet) framework, which is able to effectively integrate information of various orders, extract knowledge from high-order neighbors, and provide meaningful and interpretable results by identifying influential compact graph structures. In particular, MaGNet consists of two components: an estimation model for the latent representation of complex relationships under graph topology, and an interpretation model that identifies influential nodes, edges, and node features. Theoretically, we establish the generalization error bound for MaGNet via empirical Rademacher complexity, and demonstrate its power to represent layer-wise neighborhood mixing. We conduct comprehensive numerical studies using simulated data to demonstrate the superior performance of MaGNet in comparison to several state-of-the-art alternatives. Furthermore, we apply MaGNet to a real-world case study aimed at extracting task-critical information from brain activity data, thereby highlighting its effectiveness in advancing scientific research.
}

\vspace{2mm}


\baselineskip=23pt

\section{Introduction}

Graph-structured data is ubiquitous throughout the natural and social sciences, from brain networks to social relationships. A graph is simply a collection of nodes representing entities such as people, genes, and brain regions, along with a set of edges representing interactions between pairs of nodes. 
By representing such interconnected entities as graphs, it is feasible to leverage their geometric topology to study statistical relationships among nodes using network-based frameworks. 
Among graph representation methods, the family of graph neural networks (GNNs) has achieved remarkable success in real-world graph-based tasks \citep{velivckovic2023everything}. 
In general, GNNs iteratively aggregate and combine node representations within a graph, through a process called message passing, to generate a set of learned hidden representation features. 
The main neural architectures of GNNs include graph convolutional networks \citep[GCNs; ][]{kipf2016semi}, graph attention networks \citep[GATs; ][]{velivckovic2017graph}, graph transformer networks \citep[GTNs; ][]{yun2019graph}, among many other variants.

While GNNs are capable of capturing subgraph information through message passing, they can be prone to over-smoothing the learned representations when applying multiple rounds of message passing operations. This can cause models to treat all nodes uniformly, leading to node representations that converge into indistinguishable entities \citep{li2018deeper}. Additionally, over-smoothing could limit the ability to capture high-order information, which can be only aggregated through a sufficient number of message passing operations \citep{hamilton2020graph}. 
Several studies suggest that over-smoothing significantly contributes to deep GNN performance degradation \citep{bodnar2022neural}. 
To address this issue, \citep{zhang2022model} advocated for using shallow GNNs (e.g., up to three layers). However, this approach fails to capture high-order information due to insufficient message passing. Additionally, \cite{zhao2019pairnorm} and  \cite{yang2020revisiting} developed normalization layers to prevent node embeddings from becoming indistinguishable, but this increases training difficulty and limits the expressive power of GNNs.


A main premise of this paper is that to enhance the overall representation power of GNNs with statistical guarantees, we need to develop new learning mechanisms that directly incorporate and effectively combine information from neighbors at different orders. Statistically, by integrating both low-order (i.e., immediate neighbors) and high-order (i.e., neighbors beyond the immediate vicinity) information, GNNs can learn a richer and more complete representation under the graph topology. Models that follow the principle of effectively combining information from different orders are known as multi-scale GNNs, as they enable the exploration and integration of information at different levels of granularity within a graph
\citep{xu2018representation, sun2019adagcn,oono2020optimization,liu2022mgnni}. The main idea is to direct the outputs of intermediate layers to contribute to the final representation. Existing methods however struggle to effectively integrate representations of different orders in a sequential manner due to their memoryless property: at each GNN layer, node representations are updated entirely based on the current input from their immediate neighbors, without directly retaining information from previous layers. Furthermore, Multi-scale GNNs still suffer from over-smoothing issues and generally fail to capture high-order latent representations.

To address the aforementioned issues, we develop a novel \textbf{M}odel-\textbf{a}gnostic \textbf{G}raph neural \textbf{Net}work (MaGNet) framework consisting of two components: the \emph{estimation model} and the \emph{interpretation model}. The estimation model captures the complex relationship between the feature information and a target outcome under graph topology, allowing for powerful latent representation corresponding to unique-order graph information. 
The interpretation model, on the other hand, identifies a compact subgraph structure -- specifically, influential nodes and edges -- along with a small subset of node features that play a crucial role in the learned estimation model. The main advantages of the proposed framework, along with our contributions, are outlined below. First, the proposed neural architecture of the model alleviates the over-smoothing issue and thus can effectively extract knowledge from high-order neighbors. The proposed \emph{actor-critic} neural architecture effectively integrates multi-order information by resolving the memoryless issue. It adaptively combines representations from actor graph neural networks, each focused on a specific order, while a critic network evaluates the quality of the learned representations. Second, we develop an interpretation framework, formulated as an optimization task that maximizes information gain over the distribution of possible subgraph structures. This approach is model-agnostic as there is no assumption on the true statistical models or data-generating mechanisms. Third, we study the ability to integrate various-order information as well as the statistical complexity of the proposed model via an empirical Rademacher complexity. Unlike existing analyses limited to standard message passing neural networks, our results are applicable to a mix of message passing and feedforward neural networks, with message passing networks being a special case in our framework.
Furthermore, we provide a statistical generalization error bound for the MaGNet estimation model that consists of sequential deep-learning component models.

\section{Preliminaries}\label{prelim}

\subsection{Graph Structure}
In this section, we present some preliminaries and notations used throughout the paper. Let $G = (V,E)$ represent the graph, where $V$ represents the vertex set consisting of nodes $\{v_1,v_2,...,v_N\}$, and $E \in V \times V$ denotes the edge set with $(i,j)$th element $e_{ij}$. The number of total nodes in the graph is denoted by $N$. A graph can be described by a symmetric (typically sparse) adjacency matrix $A \in \{0,1\}^{N \times N}$ derived from $V$ and $E$. In this setting, $a_{ij} =0 $ indicates that the edge $e_{ij}$ is missing, whereas $a_{ij} =1$ indicates that the corresponding edge exists. There is a $T$-dimensional set of features, $X_i$, associated with each node, $v_i$ so that the entire feature set is denoted as $X \in \mathbb{R}^{N \times T}$. Suppose we have observed $n$ graph instances, each consisting of a fixed graph structure but with different node features. Let $G_i$ denote the $i$th instance of a graph, where $i \in {1, 2, \dots, n}$. While our approach can be used for predictive models in general, here we focus on classification problems, where the objective is to assign a binary label $s \in \{-1,1\}$ to each graph instance. 

\subsection{Neural Message Passing}

The basic graph neural network (GNN) model can be motivated in a variety of ways. The same fundamental GNN model has been derived as a generalization of convolutions to non-Euclidean data \citep{bodnar2022neural}, and as a differentiable variant of belief propagation \citep{dabkowski2017real}, as well as by analogy to classic graph isomorphism tests \citep{graham2019equivariant}. Regardless of the motivation, the defining feature of a GNN is that it uses a form of neural message passing in which vector messages are exchanged between nodes and updated using neural networks \citep{abu2019mixhop}. During each round of message passing in a GNN, a hidden embedding corresponding to each node $v \in \mathcal{V}$, denoted as $H_v^{(k)}$ for the $k$th layer where $k=1,...,K$, is updated according to information aggregated from $v$'s graph neighborhood $\mathcal{N}(v)$. This message passing update can be expressed as follows:
\$
H_v^{(k+1)} & ={f_{\text{update}}}^{(k)}\left(H_v^{(k)}, {f_{\text{agg}}}^{(k)}\left(\left\{H_u^{(k)}, \forall u \in \mathcal{N}(v)\right\}\right)\right) \\
& ={f_{\text{update}}}^{(k)}\left(H_v^{(k)}, {M}_{\mathcal{N}(v)}^{(k)}\right),
\$
where $f_{\text{update}}$ and $f_{\text{agg}}$ are the update and aggregate functions, which are arbitrary differentiable functions (here, neural networks). The term $M_{\mathcal{N}(v)}$ is the ``message'' that is aggregated from $v$'s graph neighborhood $\mathcal{N}(v)$. We use superscripts to distinguish the embeddings and functions at different rounds of message passing. At each round of message passing, the aggregate function takes as input the set of embeddings of the nodes in $v$'s graph neighborhood $\mathcal{N}(v)$ and generates a message $M_{\mathcal{N}(v)}^{(k)}$ based on this aggregated neighborhood information. The update function  then combines the message $M_{\mathcal{N}(v)}^{(k)}$ with the previous embedding $H_v^{(k-1)}$ of node $v$ to generate the updated embedding $H_v^{(k)}$.

\subsection{Graph Convolutional Network (GCN)}

 Let $\tilde{D}$ be the degree matrix corresponding to the augmented adjacency matrix $\tilde{A}=A+I$ with $\tilde{D}_{ii} = \sum^{N}_{j=1}\tilde{A}_{ij}$. The hidden graph representation of nodes with two graph convolutional layers \citep{kipf2016semi} can be formulated in a matrix form: 
\#
H = \widetilde{\mathcal{L}} \ \text{ReLU}(\widetilde{\mathcal{L}}XW^{(0)})W^{(1)},
\label{gcn}
\#
where $H \in \mathbb{R}^{N \times T^{(1)}}$ is the final embedding matrix of nodes and $T^{(1)}$ is the dimension of the node hidden representation (embedding). The graph Laplacian is defined as $\widetilde{\mathcal{L}}=D^{-\frac{1}{2}} \tilde{A} D^{-\frac{1}{2}}$. In addition, the weight matrix $W^{(0)} \in \mathbb{R}^{T\times T^{(0)}}$ is the input-to-hidden weight matrix for a hidden layer with $T^{(0)}$ feature maps, and $W^{(1)} \in \mathbb{R}^{T^{(0)} \times T^{(1)}}$ is the hidden-to-output weight matrix. Here we consider the two-layer case that aims to simplify the notation; the above definition can be easily extended to $k$ graph convolutional layers with $k>2$.

\section{Estimation Model}
\label{est_mod}

In this section, we introduce a novel graph neural network, which aims to represent feature information and capture it relationship to an outcome of interest. To achieve this goal, it effectively integrates both low-order and high-order neighbor node information to form a powerful latent representation. Here, the low-order information refers to information aggregated from the local neighbors of a node, while the high-order information means the messages aggregated beyond just the immediate/close neighbors, capturing the global graph information.

To characterize the feasibility and effectiveness of capturing various-order information rigorously, we first introduce a generalized 2-order $\Delta$-representer, which is defined by \citet{abu2019mixhop}, i.e., the corresponding $K$-order counterpart for $K \geq 3$ as follows.

\begin{definition}
 \label{mixing}
Given a graph neural network, $\Delta(K)$-representer represents $K$-order node neighbor information for $K \in \mathbb{N}$, e.g., there exists a real-valued vector $\nu = (\nu_1,\nu_2,...,\nu_K)$ and an injective (one-to-one) mapping function $g(\cdot)$, such that the output embedding of this graph neural network can be represented as  
\$
 g\left(\sum_{k=0}^K \nu_k \cdot \mathcal{L}^k X\right)  := \Delta(K),
\$
for any type of graph Laplacian $\mathcal{L}$ operation and input node feature matrix $X$. 
\end{definition}

Learning such a representer enables GNNs to capture feature differences among $K$-order node neighbor's information. 
When a candidate GNN model learns the $\Delta(K)$-representer, it effectively captures the $K$-order neighborhood information in the hidden representation. 

In GCNs, the graph representation is obtained through interactions of neighboring nodes during multiple rounds of learned message passing. Ideally, one could consider a deep architecture via stacking $K$ GCN layers in order to learn a $\Delta(K)$-representer. However, most of the existing GCN models employ shallow architectures, typically utilizing only second- or third-order information \citep{zhang2020deep}. The reason behind this limitation is two-fold. First, when repeatedly applying Laplacian smoothing, GCNs may mix node features from different clusters, rendering them indistinguishable. This phenomenon is known as the \textit{over-smoothing} issue \citep{li2021training}. Second, most GCNs are built upon a feedforward mechanism and suffer from the \textit{memoryless} problem. After each layer operation, the representation learned from the current layer modifies the representation produced from the previous layers. As a result, there is no explicit memory mechanism. In other words, the \textit{over-smoothing} issue creates difficulties in capturing high-order information, while the \textit{memoryless} issue leads to a loss of lower-order information. Theoretically, under Definition \ref{mixing}, the GCN models with over-smoothing or memoryless issues cannot learn the $\Delta(K)$-representer. A more detailed discussion is provided in the Appendix.

\subsection{Actor-Critic Graph Neural Network}

In this section, we describe our actor-critic graph neural network, which is designed to effectively aggregate different levels of node-neighbor information to obtain a powerful graph embedding. In this dual neural network structure, the actor graph neural network aims to capture the hidden representation for each order of node-neighbor information, while the critic neural network plays the role of evaluating the quality of the hidden representation learned by the actor network. To this end, we perform a fusion operation to integrate the representations from individual actor networks using the corresponding quality scores as weights. This framework resolves the over-smoothing and memoryless issues, ensuring a properly learned $\Delta(K)$-representer.

In contrast to GCNs, we adopt the
simple weighted sum aggregator and abandon the nonlinear transformation. As a result, the graph convolution operation in our actor graph neural network is defined as: 
\#
H^{(k)} =  (\mathcal{L})^{k}XW,
\label{light}
\#
where the graph Laplacian $\mathcal{L}=D^{-\frac{1}{2}} A D^{-\frac{1}{2}}$, and $W$ is the weight matrix, which can be fixed as identity matrix in every message passing round. We argue that the majority of the benefit arises from the local averaging of neighboring features. This is because unlike multi-dimensional image data, the vectorized temporal signal does not require many nonlinear layers to capture the information. Furthermore, the nonlinear feature transformation in GNNs is useful but not critical \citep{wu2019simplifying}. By removing the nonlinear activation, our graph convolution layer achieves slower
convergence in certain values for embedding vectors, thus alleviating the over-smoothing issue. Additionally, abandoning the nonlinear feature transformation operation greatly improves computation. 

It is also worth noting that in \eqref{light}, we aggregate only the connected neighbors without integrating the target node itself. That is, the graph Laplacian is based on the adjacency matrix $A$ instead of its augmented counterpart $\tilde{A}$. The fusion operation in our model, to be discussed later, essentially captures a similar effect as ``self-connection'' with adaptivity. This \emph{partial self-connection} mitigates over-smoothing issues. In the Appendix, we provide a discussion on this theoretical investigation of the partial self-connection in Theorem S.1. Notably, this distinguishes our model from the closely related GNN works, e.g., \citep{kipf2016semi,sun2019adagcn, wu2021self}, that aggregate extended
neighbors and need to handle the self-connection explicitly without flexibility.

Further, in contrast to node classification, our interest lies in graph classification tasks. Therefore, based on the learned node representation $H^{(k)}$, we can apply a graph pooling operation to summarize the graph embedding from $H^{(k)}$. The goal of the graph pooling operation is to aggregate information across the entire graph to produce a single, fixed-size representation. Specifically, we could use a simple average graph pooling operation to obtain graph embeddings with a compact node representation: 
$
\bar{H}^{(k)} =\frac{1}{N} \mathbf{1}_N H^{(k)}.
$
See the Appendix for other types of graph pooling operation, and \cite{hamilton2020graph} for a more comprehensive review. 

\begin{figure}[t]
\captionsetup{font=normalsize}
  \captionsetup{width=.9\linewidth}
	\centering	\scalebox{0.6}[0.6]{\includegraphics{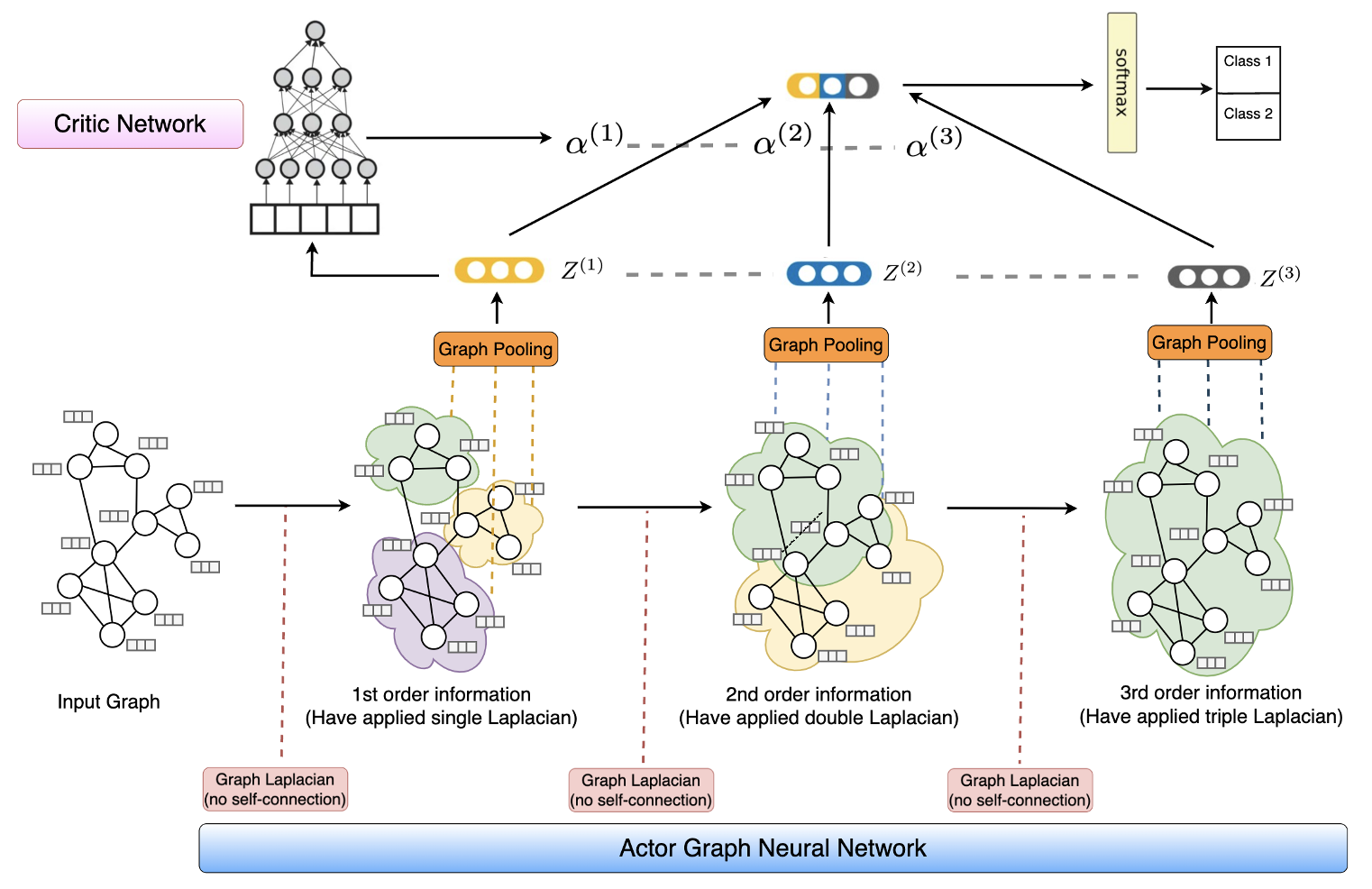}}
	\caption{An illustrative example of $3$-layers neural architecture of the actor-critic graph neural network.}\label{fig:archi1}
\end{figure}
As previously stated, our primary objective is to aggregate mixed-order information. To this end, we propose a fusion operation to completely preserve the information from various-order neighbors. Specifically, we can regard the graph embedding $\bar{H}^{(k)}$ from $k$th round of message passing as the output of the $l$th order information summarization, denoted as $\bar{H}^{(k)}$. From the perspective of meta algorithms, the embedding $\bar{H}^{(k)}$ can be regarded as the graph embedding learned from the $k$th actor graph neural network. Then the ultimate graph embedding can be weighted combined from the various order graph embeddings, i.e.,
\#
\widetilde{H} = \sum^{K}_{k=1} \alpha^{(k)} \bar{H}^{(k)},
\label{agg_embedd}
\#
where $\alpha^{(k)}$ is the fusion weight 
corresponding to the quality (or importance) of the $k$th order knowledge for $k=1,...,K$. This fusion operation can be understood as the ensemble of multiple single actor networks. i.e., the actor networks for generating graph embedding $\bar{H}^{(k)}$. Thus, the fusion operation naturally combines the unique characteristics of different single learners with different order information. Moreover, as we discussed previously, the fusion operation intrinsically captures the so-called partial self-connection effect and thus relaxes over-smoothing issues. Theorem S.2 in the Appendix provides another view of the partial self-connection effect from the perspective of the graph spectral analysis. It shows that our model is more capable of mitigating the over-smoothing issue in comparison to GCNs.

To determine the fusion weights $\alpha^{(k)}$, we introduce a critic network, $f_{\text{critic}}: \bar{H}^{(k)} \mapsto \Delta_{\{-1,1\}}$ for a probability simplex $\Delta_{\{-1,1\}}$, which takes the graph embedding vector as input and output the binary probability logits. Generally, this critic network could be any proper parametric or non-parametric classification model including softmax (logistic) regression, random forest, feedforward neural network (multilayer perception, MLP), and so on. Without loss of generality, we choose the critic network as a class of the softmax regression in the following for illustration purposes.

The critic network plays a role in evaluating the quality of the graph embedding $\bar{H}^{(k)}$ in a bias induction way. That is 
\$
\alpha^{(k)} = \frac{1}{2}\log\left(\frac{1-\epsilon^{(k)}}{\epsilon^{(k)}}\right) ,
\$
where the function $\text{logit}(x) = \log(x/(1-x))$ and the error rate $\epsilon^{(k)}$ is defined as
\#
\epsilon^{(k)} = \sum^{n}_{i=1}\beta^{(k)}_i \mathds{1}\left\{s_i \neq \argmax_{\{s=-1,1\}}\text{softmax}(\bar{H}^{(k)}_i)\right\}/\sum^{n}_{i=1}\beta^{(k)}_i.
\label{iter_err}
\#
Here $\bar{H}^{(k)}_i$ denotes the graph embedding for the $i$th graph sample, $\argmax_{\{s=-1,1\}}(x)$ is the operator taking the maximum element in the two-dimensional vector $x$, and the coefficient $\beta^{(k)}_i$ denotes the weight of the $i$th graph sample. Intuitively, $\epsilon^{(k)}$ can be understood as the weighted classification error rate for the $k$th critic network, i.e. $\text{softmax}(\bar{H}^{(k)})$. For $i=1,...n$, we adjust the graph sample weights $\beta^{(k)}_i$ sequentially from the $k$th to the $(k+1)$th step by following the updating rule:
$
\beta^{(k+1)}_i  \propto \beta^{(k)}_i e^{\mathds{1}\{s_i \neq \argmax_{\{s=-1,1\}}\text{softmax}(\bar{H}^{(k)}_i)\} \cdot \alpha^{(k)}} . 
$
This update rule intentionally pays more attention to the misclassified graph samples with potentially insufficient representation power and increases their weights when training the next single actor network. Notably,  instead of enhancing the nonlinear representation for embedding vectors, e.g., $\bar{H}^{(k)}_i$ during the phase of the fusion, we might opt to perform nonlinear transformation on the graph embedding $\widetilde{H}$. This helps to relax the training difficulties well, in contrast to the existing \citep{ivanov2021boost,  sun2019adagcn} where the nonlinear transformation is performed layer-wisely. 

The embedding of the ultimate graph $\widetilde{H}$ in \eqref{agg_embedd} combines information adaptively from the $1$st to the $K$th order node neighbors, using adaptive weights. Furthermore, the sequential updates of single actor networks maintain the same learning pattern as standard GCNs, wherein the message passing for the $(k+1)$th hop directly succeeds the message passing for the $k$th hop. 
In this manner, our actor-critic graph neural network maintains most of the desirable properties found in standard GCNs, including invariance to graph isomorphism \citep{xu2018powerful} and effective relational representation \citep{wu2020comprehensive}.

Ultimately, we establish the classification model and the rule for prediction based on the graph embedding $\widetilde{H}_i$ corresponding to the $i$th graph sample, 
\#
p(\cdot|\widetilde{H}_i) = \text{softmax}(g(\widetilde{H}_i)).
\label{final_mod}
\#
where $g(\cdot)$ is an arbitrary function to perform the linear/nonlinear (or even identity) transformation mapping. The
equation returns the classification logits. To streamline the notation throughout the paper, we use $p_{\widehat{\theta}}(\cdot)$ to represent a trained actor-critic graph neural network model. To have a better understanding of the proposed actor-critic neural network.

Finally, we note that training the model in \eqref{final_mod} is a well-studied convex optimization problem. It can be performed using efficient second-order methods or stochastic gradient descent (SGD) \citep{bottou2010large}. As long as the graph connectivity pattern remains sufficiently sparse, SGD can naturally scale to handle very large graph sizes. 
Furthermore, we ensure the neural network architecture's consistency by training the layers sequentially, following the order of node neighbors' message passing. This sequential training approach enables us to utilize the trained parameters from the previous training to initialize the current model training, which effectively reduces computational costs.

\section{Interpretation Model}\label{interp_model}

Although the estimation model provides strong representation power to capture the complex relationship between the outcome of interests and features, understanding the rationale behind its predictions can be quite challenging. In this section, we present a practically useful interpretation framework designed to uncover the reasoning behind the ``black-box'' estimation model.

To bridge the gap between estimation and interpretation, we first observe that our estimation model extracts feature information from various-order node neighbors as well as graph topology to output the hidden graph representation for predictions. This suggests that the prediction made by the estimation model, i.e., $\widehat{s} = \argmax_{\{s=-1,1\}} p_{\widehat{\theta}}(\cdot)$ in \eqref{final_mod}, is determined by the adjacency matrix $A$ and the node feature information $X$. Formally, to comprehend the model mechanism and provide explanations, the problem is transformed into identifying important subgraphs, denoted as $G_{sub} \subseteq G$ with a corresponding adjacency matrix $A_{sub}$, along with a small subset of the node feature $X_{sub}$ in full dimension. We first focus on the identification of influential subgraphs by assuming $X_{sub}$ has been obtained and then discuss how to perform node feature selection simultaneously with subgraph identifications.

We adapt the principle of information gain, which was first introduced in the context of decision trees \citep{larose2014discovering}, into our framework. In particular, we formulate an optimization framework for influential subgraph identification. Our goal is to maximize the information gain with respect to subgraph candidates $G_{\text{sub}}$:
\#
\underset{G_{sub}}{\operatorname{argmax}} \, \text{IG}\left(p_{\widehat{\theta}}, G_{sub}\right)=\eta(p_{\widehat{\theta}})-\eta\left(p_{\widehat{\theta}} \mid G_{sub}, X_{sub} \right),
\label{EI}
\#
where $\eta(\cdot)$ and $\eta(\cdot|\cdot)$ denote the entropy and conditional entropy respectively. 

Essentially, information gain can quantify the change in prediction probability between the full model $p_{\widehat{\theta}}(\cdot)$ and the one constrained to the subgraph $G_{sub}$ and the subset node feature $X_{sub}$ \cite{ying2019gnnexplainer}.  For example, if removing edge $e_{ij}$, i.e., the $(i,j)$th element in the adjacency matrix $A$, from the full graph $G$ significantly decreases the prediction probability, then this edge is influential and should be included in the subgraph $G_{sub}$. Conversely, if the edge $e_{ij}$ is deemed redundant for prediction by the learned estimation model, it should be excluded.

Examining the right-hand side of \eqref{EI}, we can easily observe that the entropy term $\eta(p_{\widehat{\theta}})$ remains constant since the parameters $\widehat{\theta}$ are fixed for an estimated model. Consequently, the objective of minimizing information gain in \eqref{EI} is equivalent to maximizing the conditional entropy $\eta\left(p_{\widehat{\theta}} \mid G_{sub}, X_{sub} \right)$. Nevertheless, directly optimizing the above objective function is intractable, as there are $2^{|V|}$ candidates for the subgraph $G_{sub}$. To address this issue, we consider a relaxation by assuming that the subgraph is a Gilbert random graph \citep{reitzner2017limit}. This way, the selection of edges from the original input graph $G$ are conditionally independent of each other and follow a probability distribution. In detail, the edge $e_{ij}$ is a binary variable indicating whether the edge is selected, with $e_{ij}=1$ if selected and $0$ otherwise. Therefore, the graph $G_{sub}$ is a random graph with probability
$
P(G_{sub}) = \Pi_{i,j \in N}P(e_{ij})
$.
 A straightforward instantiation of $P(e_{ij})$ is the Bernoulli distribution $e_{ij} \sim \text{Bern}(\mu_{ij})$, where $\mu_{ij}$ is the first moment. In particular, we can rewrite the parametrized objective as: 
\#
\underset{G_{sub}}{\text{Minimize}}\; \eta\left(p_{\widehat{\theta}} \mid G_{sub}, X_{sub} \right) = \underset{G_{sub}(\mu)}{\text{Minimize}} \; \mathbb{E}_{G_{sub}(\mu)}[\eta\left(p_{\widehat{\theta}} \mid G_{sub}, X_{sub} \right)] , 
\label{lower}
\#
where $G_{sub}(\mu)$ is the parametrized random subgraph. Due to the discrete nature of the subgraph $G_{sub}(\mu)$, the objective function is non-smooth, making optimization challenging and unstable. To address this issue, we further leverage a continuous approximation for the binary sampling process \citep{maddison2016concrete,luo2020parameterized}. Let  $\epsilon$ be a uniform random variable, i.e., $\epsilon \sim \text{Unif}(0,1)$, and the real-valued parameters $\psi_{ij} \in \Psi$, and a temperature parameter $\omega \in \mathbb{R}^{+}$, then a sample of the binary edge $e_{ij}$ can be approximated by a sigmoid mapping:
\$
\widetilde{e}_{ij} = \text{sigmoid}\left(\frac{\log(\epsilon) - \log(1-\epsilon) + \psi_{ij}}{\omega}\right).
\$
We denote $\widetilde{G}_{sub}(\Psi)$ as the continuous relaxation counterpart of the subgraph, with the $(i,j)$th element of the adjacency matrix being $\widetilde{e}_{ij}$. Interestingly, the temperature parameter $\omega$ can describe the relationship between $\widetilde{G}_{sub}(\Psi)$ and $G_{sub}(\mu)$. We observe that as $\omega \rightarrow 0$, the approximated edge $\widetilde{e}_{ij}$ converges to the edge $e_{ij}$, with the probability mass function, 
$
\lim_{\omega \rightarrow 0}P(\widetilde{e}_{ij} = 1) = \frac{\exp(\psi_{ij})}{1+\exp(\psi_{ij})}. 
$
Recall that the edge $e_{ij}$ follows a Bernoulli distribution with mean $\mu_{ij}$. If we reparameterize $\psi_{ij}$ such that
$
\psi_{ij} = \log\left(\frac{\mu_{ij}}{1-\mu_{ij}}\right)$,
it achieves asymptotical consistency of the approximated subgraph by following the limiting theory \citep{paulus2020gradient}, i.e., $\lim_{\omega \rightarrow 0} \widetilde{G}_{sub}(\Psi) = {G}_{sub}(\mu)$. This supports the feasibility of applying continuous relaxation to the binary distribution.

Unlike the objective function in \eqref{lower}, which is induced by the discrete original subgraph, the objective function becomes smooth under the edge continuous approximation and can be easily optimized using gradient-based methods. In other words, the gradient of the continuous edge approximation $\widetilde{e}_{ij}$ with respect to the parameters $\psi_{ij}$ is computable. More importantly, the sampling randomness towards the subgraph is absorbed into a uniform random variable $\epsilon$ peel-off from the parameterized binary Bernoulli distribution, which greatly relaxes the complexity of the sampling processing. 

In this manner, the objective function in \eqref{lower} can be reformulated as
\$
\underset{\Psi}{\text{Minimize}} \; \mathbb{E}_{\epsilon \sim \text{Unif}(0,1)}[\eta\left(p_{\widehat{\theta}} \mid G_{sub}(\Psi), X_{sub} \right)]. 
\$
However, solving the conditional entropy is still computationally expensive. To avoid this issue, we follow \cite{kipf2018neural} to minimize a cross-entropy as the objective function. We should note that the conditional
entropy is upper bounded by cross-entropy, which validates the possibility to minimize the cross-entropy objective. In particular, the empirical objective becomes
\$
\underset{\Psi}{\text{Minimize}} \; \frac{1}{n}\sum_{i=1}^{n}\sum_{s \in \{-1,1\}} p_{\widehat{\theta}}(s_i=s|X_{sub}(i)) \log p_{\widehat{\theta}}(s_i=s|G_{sub}(\Psi), X_{sub}(i)),
\$
where $n$ is the sampling size and $p_{\widehat{\theta}}(s_i=s|G_{sub}(\Psi),X_{sub}(i))$ denotes the classification logits conditional on the subgraph $G_{sub}(\Psi)$ and the subset feature $X_{sub}(i)$ of the $i$th graph sample.

So far, we have implicitly assumed that the subset feature $X_{sub}$ is known. This is not the case in practice. In the context where the subset feature $X_{sub}$ is not given, the main challenges are: 1) identifying the subset is unknown; and 2) the fact that integrating this feature selection into the developed subgraph identification optimization framework is not trivial.
Motivated by the great success of self-supervised techniques in large neural language models \citep{devlin2018bert}, we propose to use a ``masking'' approach to convert the feature subsetting problem into an optimization problem that can be naturally combined with subgraph identification. Specifically, we define a binary vector $\mathcal{B} \in \{0,1\}^{T}$ which holds the same dimension as the raw node feature. For each node $v_i$ and its raw node feature $X_i$, $i=1,...,N$, we multiply the raw feature with the binary vector $\mathcal{B}$ to obtain $X_i \odot \mathcal{B}$, where $\odot$ is the Hadamard product. Intuitively, the vector $\mathcal{B}$ converts the value in some dimension of the node feature to $0$. This aligns with the rationale that if a particular feature is not important, the corresponding weights in the neural network weight matrix take values close to $0$. 
In terms of the principle of information gain, this type of masking does not significantly decrease the probability of the prediction or alter the information gain. 

The binary vector $\mathcal{B}$ is non-smooth; we also consider a continuous relaxation for the vector by leveraging a sigmoid mapping, so that the feature selection procedure becomes a smooth optimization problem, i.e., 
\$
X \odot \text{sigmoid}(\widetilde{\mathcal{B}}),
\$
where $\widetilde{\mathcal{B}} \in \mathbb{R}^{T}$ is a real-valued vector and the $\text{sigmoid}(\widetilde{\mathcal{B}})$ is applied to each row of $X$. Next, we remove the low values in $\widetilde{\mathcal{B}}$ through thresholding to arrive at the feature subsetting. 

The subgraph identification and the feature selection can be naturally integrated into a single minimization problem:
\$
\underset{\Psi,\widetilde{\mathcal{B}}}{\min} \; \frac{1}{n}\sum_{i=1}^{n}\sum_{s \in \{-1,1\}} p_{\widehat{\theta}}(s_i=s|X(i) \odot \text{sigmoid}(\widetilde{\mathcal{B}})) \log p_{\widehat{\theta}}(s_i=s|G_{sub}(\Psi), X(i) \odot \text{sigmoid}(\widetilde{\mathcal{B}})),
\$
which forms a unified optimization framework. This unified optimization framework allows for the simultaneous identification of influential subgraphs and important node features, leading to a more interpretable and efficient model. The resulting optimization problem can be solved using gradient-based techniques.


\section{Theory}
\label{theory_sec}
In this section, we present the main theoretical results of the estimation model. First, we demonstrate that in contrast to standard GCNs, our approach is capable of effectively representing the feature differences among various-order neighbors. Second, we study the capacity of the actor graph neural network in terms of empirical Rademacher complexity. The derived bound is tight through careful analysis of both the lower and upper bounds. Furthermore, we provide a probabilistic upper bound on the generalization error of the actor-critic graph neural network which is calibrated in the fusion algorithm.

\begin{thm}
\label{able_k}
The MaGNet actor-critic graph neural network is capable of learning a $\Delta(K)$-representer, which means it can sufficiently and effectively capture $K$-order node neighbor information.
\end{thm}

Theorem \ref{able_k} demonstrates that our estimation model can learn various-order information. This ensures the capability of the proposed estimation model on high-order message passing, where nodes receive latent representations from their $1$-order neighbors as well as further $K$-order neighbors at the information aggregation step. In contrast, the existing GCNs are not capable of representing this class of operations, even when stacked over multiple layers. Please see further justification for this statement in Section B of the Appendix. To establish the bounds on generalization errors, we make the following technical assumptions.

\begin{assumption}
\label{input_bound}
The feature vector of any graph is contained in a $L_2$-ball with radius $\widetilde{c}$. Specifically, the $L_2$ norm of the feature vector $\left\|X_i\right\|_2 \leq \widetilde{c}$ for all $i=1,...,N$ and some constant $\widetilde{c}>0$.
\end{assumption}

\begin{assumption}
\label{mat_bound}
Any weight matrix in the estimation model satisfies that 
 \$
\|w^{(l)}_{\text{MLP}}\|_{F} \leq c_2, \|w^{(l_0)}_{\text{MLP}}\|_{2} \leq c_1, \|W\|_{F} \leq c_0,
 \$
 with some constant $c_0, c_1, c_2 > 0$ and the Frobenius norm $\|\cdot\|_{F}$, where $w^{(l)}_{\text{MLP}}$ is the weight matrix in $l$th layer of MLP for $l=1,...,l_0-1$, and $w^{(l_0)}_{\text{MLP}}$ is the weight vector in the last layer of MLP.
\end{assumption}

\begin{assumption}
\label{lap_bound}
The maximum number of elements in the graph Laplacian matrix is bounded above by, i.e., $\max_{i \in [N]}\max_{j \in [N]}|\mathcal{L}_{ij}| \leq c_{\mathcal{L}}$.
\end{assumption}

\begin{assumption}
\label{neibor_assume}
We only consider the undirected, no loops
, and no multi-edges graphs, and the number of node neighbors $|\mathcal{N}(v_i)|$ for all node $v_i \in V$ is equal to some constant $q \in \mathbb{N}^{+}$. 
\end{assumption}

\begin{assumption}
\label{layer_assume}
The maximum hidden dimension across all neural network layers is $h$.
\end{assumption}

The above assumptions are common in the (graph) neural network literature. 
Assumptions \ref{input_bound}-\ref{mat_bound} impose norm constraints on the parameters and input feature, making the model class fall into a compact metric space \citep{liao2020pac}. In general, Assumption \ref{input_bound} does not require a specific data distribution for the feature vector. It holds as long as the feature vector has a bounded $L_2$-norm, regardless of its distribution. For example, feature vectors following the truncated Gaussian distribution, uniform distribution, logarithmic distribution, and autoregressive distribution within the $L_2$-ball all satisfy Assumption \ref{input_bound}. Assumption \ref{lap_bound} is a standard assumption to control the intensity of the graph Laplacian in GNN literature \citep{hamilton2020graph}. Assumption \ref{neibor_assume} requires us to focus on homogeneous graphs \citep{liao2020pac,lv2021generalization}. Assumption \ref{layer_assume} is a standard assumption in bounding the width of neural network layers. 

We first present our result on bounding the Rademacher complexity of the model class $\mathcal{F}_{c_0,c_1,c_2}$, which is the estimation model part before and up to the step that produces $H^{(K)}$. For $i_0$th graph sample, formally, we define our estimation model class $\mathcal{F}_{c_0,c_1,c_2}$ in the setting of $K=3$ and $l_0=2$ without loss of generality: 
\$
\mathcal{F}_{c_0,c_1,c_2} := \bigg\{ & f(X(i_0)) =  \sigma\Bigg(\sum^{d_1}_{q=1} {w^{(2)}_{\text{MLP}}}_{q}\sigma\Bigg(\sum^{k}_{t=1}{w^{(1)}_{\text{MLP}}}_{tq} \frac{1}{N}\sum^{N}_{m=1}  \sum_{i=1}^N\mathcal{L}_{m i}\sum_{v=1}^N\mathcal{L}_{i v} \notag \\
& \qquad \qquad \qquad  \times 
  \sum_{j \in \mathcal{N}(v)}\mathcal{L}_{v j}\left\langle X(i_0)_j, \mathbf{w}_t\right\rangle \Bigg)\Bigg), \quad i_0 \in [n], \notag \\
  & \quad \|w^{(1)}_{\text{MLP}}\|_{F} \leq c_2,  \|w^{(2)}_{\text{MLP}}\|_{2} \leq c_1, \|W\|_{F} \leq c_0 \bigg\},
\$
where $\sigma(\cdot)$ is some activation function, $w^{(1)}_{\text{MLP}}$ and $w^{(2)}_{\text{MLP}}$ is the weight matrix and vector for the first and second layer of the MLP for critic network, respectively. The $\mathbf{w}_t$ is the $t$th column of the weight matrix $W$. Note that we use this particular setting as an example of the model class for simplifying the expression. The following theoretical results hold for the general case of $K$ and $l_0$.

\begin{definition}
Given the input node feature matrix $\{X(i)\}^{n}_{i=1}$ and the model class of the actor-critic graph neural network $\mathcal{F}_{c_0,c_1,c_2}$, the empirical Rademacher complexity of $\mathcal{F}_{c_0,c_1,c_2}$ is definied as 
\$
\widehat{\mathcal{R}}(\mathcal{F}_{c_0,c_1,c_2} ):=\mathbb{E}_\epsilon\left[\frac{1}{n} \sup _{f \in \mathcal{F}_{c_0,c_1,c_2}}\bigg|\sum_{j=1}^n \epsilon_j f\left(X(j)\right)\bigg| X(1), X(2), \ldots, X(n)\right],
\$
where $\left\{\epsilon_i\right\}_{i=1}^n$ is an i.i.d. family of Rademacher variables, independent of $\{X(i)\}^{n}_{i=1}$.
\end{definition}

\begin{thm}
\label{rad_mea}
Under Assumptions \ref{input_bound}-\ref{layer_assume}, the empirical Rademacher complexity is bounded by 
\$
\widehat{\mathcal{R}}(\mathcal{F}_{c_0,c_1,c_2}) \leq &  \frac{3(L_0)^{l_0}c_0c_1c_2c^{K-1}_{\mathcal{L}}\widetilde{c}(K+l_0)h^{1.5}q^{K+0.5}}{2\sqrt{n}} |\lambda_{\max}({\mathcal{L}})|; \qquad   \qquad  \qquad \; \text{Upper Bound} \\
\widehat{\mathcal{R}}(\mathcal{F}_{c_0,c_1,c_2}) \geq &
\frac{(L_0)^{l_0}c_0c_1c_2(\min_{m,i \in [N]}\mathcal{L}_{m i})^{K-1}\widetilde{c}(K+l_0)h^{1.5}q^{K}}{5\sqrt{n}} |\lambda_{\min}({\mathcal{L}})|; \qquad  \; \text{Lower Bound}
 \$
where $L_0$ is the Lipschitz constant for activation function $\sigma(\cdot)$ in the critic network, and $\lambda_{\min}(\mathcal{L})$ and $\lambda_{\max}(\mathcal{L})$ are the finite minimum and maximum absolute eigenvalue of graph Laplacian $\mathcal{L}$.
\end{thm}

Theorem \ref{rad_mea} demonstrates that our derived upper bound is tight up to some constants when comparing it to the lower bound. Theorem \ref{rad_mea} indicates that the upper bound of $\widehat{\mathcal{R}}(\mathcal{F}_{c_0,c_1,c_2} )$ depends on the number of graph instances, the node degree of the graph, and the graph convolution filter, and also the maximum width of the neural networks. Interestingly, the above bound is independent of the maximum number of nodes, $N$, for traditional regular graphs. Note that while \cite{esser2021learning} also examine the relation between the graph information and the feature information, their bounds are not directly comparable to our theoretical results. \cite{lv2021generalization} also establishes the Rademacher complexity bound; however, the focus is only on node classification tasks and graph neural networks with one hidden layer.

Applying our results in empirical Rademacher complexity $\widehat{\mathcal{R}}(\mathcal{F}_{c_0,c_1,c_2} )$ to generalization analysis, we now state the fundamental result of the generalization bound of the estimation model. We denote $conv(\mathcal{F}_{c_0,c_1,c_2} )$ as the closed convex hull of $\mathcal{F}_{c_0,c_1,c_2} $. That is, $conv(\mathcal{F}_{c_0,c_1,c_2} )$ consists of all functions that are pointwise limits of convex combinations of functions from $\mathcal{F}_{c_0,c_1,c_2}$:
\$
\operatorname{conv}(\mathcal{F}_{c_0,c_1,c_2} ):= \Big\{& f: \forall x, f(x)=\lim_{K \rightarrow \infty} f_{K }(x), f_{K}=\sum^{K }_{k=1} w_k f_k,  \\ 
& \qquad \qquad \sum_{k=1}^K  w_k=1,
 f_k \in \mathcal{F}_{c_0,c_1,c_2} , K \geq 1\Big\} .
\$
Obviously, we can observe that the combination in \eqref{agg_embedd} belongs to $conv(\mathcal{F}_{c_0,c_1,c_2} )$. Next, we present the probabilistic generalization error for the estimation model.

\begin{thm}
\label{generalerror}
Under Assumptions \ref{input_bound}-\ref{layer_assume}, given $\widehat{s}$ as the predicted label from an $K$-layers actor-critic graph neural network with true label $s_0$, then the probabilistic upper bound of the generalization error 
\$
P& \left(\widehat{s}s_0 \leq 0 \right) \leq \mathcal{O} \vast( \prod_{k=1}^{K} \Bigg\{  \underbrace{ \sqrt{\epsilon^{(k)}\left(1-\epsilon^{(k)}\right)} + \bigg(\frac{\log \log _2\big(2\big(\log \prod_{k=1}^{K} \sqrt{\frac{1-\epsilon^{(k)}}{\epsilon^{(k)}}} \vee 1\big)\big)}{n}\bigg)^{0.5}}_{\text{fusion estimation bias}} \\
+ &      \underbrace{\sqrt{\frac{1}{2 n} \log \frac{2}{\delta}}}_{\text{intrinsic uncertainty}} +
\underbrace{\frac{(L_0)^{l_0}c_0c_1c_2c^{K-1}_{\mathcal{L}}\widetilde{c}(K+l_0)h^{1.5}q^{K+0.5}}{\sqrt{n}} |\lambda_{\max}({\mathcal{L}})|
\Bigg(\log \prod_{k=1}^K \sqrt{\frac{1-\epsilon^{(k)}}{\epsilon^{(k)}}} \vee 1\Bigg)}_{\text{local complexity}}  \Bigg\} \vast),
\$
with probability at least $1-\delta$ for $\delta \in [0,1)$, where $\vee$ is a maximum operator.  
\end{thm}

Theorem \ref{generalerror} demonstrates that the generalization error of the proposed estimation model is bounded in terms of the error rate at the $k$th iteration defined in \eqref{iter_err}, i.e., $\epsilon^{(k)} \geq 1/2$ for any $k=1,...,K$. In comparison to the generalization bound on vanilla GNNs \citep{scarselli2018vapnik,garg2020generalization}, our bound is independent of the number of hidden units and the maximum number of nodes $N$ in any input graph.
For a regular graph with $q = \mathcal{O}(1)$ \citep{bollobas1998modern}, we conclude that $\lambda_{\max}(\mathcal{L}) =1$, which yields a generalization error bound of order $\mathcal{O}(1/\sqrt{n})$ that is fully independent of the number of nodes $N$.

\section{Simulation Studies}
\label{numer}

In this section, we present a comprehensive evaluation of MaGNet using synthetic datasets. 
To generate graphs, we allow the number of nodes $N$ in the graph to vary with different graph sample sizes $n$. Each node has a $p$-dimensional feature in estimation or interpretation tasks. We distinguish two categories of nodes, specifically, important nodes and non-important nodes, and we generate their features by applying two separate processes, resulting in two different settings. To establish graph structure, we calculate the correlation across the varying node features to obtain the adjacency matrix.  In all the experiments, we use a binary outcome of interest. In what follows, we illustrate the data-generating process for each setting.

\textbf{Setting 1}: For the important nodes, features are generated by following a multivariate Gaussian distribution, $\text{MVN}(0,0.1 \cdot I)$, where $I$ is an identity matrix. On the other hand, for the non-important nodes, the instance features are sampled from a uniform distribution, $\text{Unif}(0,1)$. The difference in feature generation mechanisms creates a distributional gap influencing the classification target outcome. It is important to note that the target outcome or classification rule is based solely on the features of important nodes and remains independent of those of non-important nodes. This generation process allows a good classifier to be able to separate important nodes from non-important ones. In this setting, we form a linear classification rule $\frac{\mathbf{e}^{\top}X_{V_0}\mathbf{e}}{|V_0|} + N(0,0.1) > 0$, where $\mathbf{e}$ is the column vector whose entries are all $1$'s. Here, $X_{V_0}$ is the node feature matrix associated with the important node set $V_0$. It has dimension $|V_0| \times p$, where $|\cdot|$ is the cardinality operator. 

\textbf{Setting 2}: For the important nodes, features are generated following a Gaussian process in order to introduce a dependency between the temporal features. The mean function  $m(x_t)$ for $t=1,...,p$ and $x_t \sim \text{Unif}(0,1)$. The kernel covariance function of the Gaussian process is $k\left(x_t, x_{t^{\prime}}\right)=\sigma^2\exp \left(-\frac{1}{l^2} \left|x_t-x_{t^{\prime}}\right|^2\right)$, where $l=1$ and $\sigma=1$.
In contrast, the instance features in the non-important nodes are sampled from a Gaussian process with the same mean function, but $\sigma$ is set to $2.5$ in the kernel covariance function.  In this setting, we use a nonlinear and complex classification rule as follows:
$$
\sin(\mathbf{x}\mathbf{e}_1)\cdot \cos(\mathbf{x}\mathbf{e}_2) + \mathbf{x}^{\circ 3}\mathbf{e}_3 +N(0,0.1) >0,
$$
where $\circ$ is Hadamard power, and the row vector $\mathbf{x} = \frac{\mathbf{e}^{\T}X_{|V_0| \times p}}{|V_0|}$. Moreover, $\mathbf{e}_1, \mathbf{e}_2, \mathbf{e}_3 $ are column vectors with dimension $p$. In particular, $\mathbf{e}_1 = [\underbrace{1,1,...,1}_{\floor*{p/3}},0,...,0]$, $\mathbf{e}_2 = [0,...,0,\underbrace{1,1,...,1}_{\floor*{p/3}},0,...,0]$, $\mathbf{e}_3 = [0,...,0,\underbrace{1,1,...,1}_{\floor*{p/3}}]$.

\subsection{Evaluation of the Estimation Model}\label{sec:simu_class}

In this section, we evaluate the classification accuracy of the MaGNet estimation model by comparing it against several state-of-the-art GNN approaches, including GPS Graph
Transformer Network (GPS,
\cite{rampavsek2022recipe}), Graph Transformer Network (GTN, \cite{yun2019graph}), Mean-Subtraction-Norm Graph Convolutional Network (MSGCN, \cite{yang2020revisiting}), PairNorm Graph Attention Network (PNGAT, \cite{zhao2019pairnorm}), and Approximate Personalized Propagation of Neural Predictions (APPNP, \cite{gasteiger2018predict}). The results are provided in Table \ref{class_1} and Table \ref{class_2}.

\begin{table}[H]
\centering
\scriptsize
\begin{tabular}{@{}ccccccccc@{}}
\toprule
\textbf{Sample Size} & \textbf{Important Nodes} & \textbf{Nodes} & \textbf{MaGNet} & \textbf{PNGAT} & \textbf{GTN} & \textbf{GPS} & \textbf{MSGCN} & \textbf{APPNP} \\
\midrule
$100$ & $10$ & $30$ & $\mathbf{0.761}$ & $0.715$ & $0.745$ & $0.742$ & $0.719$ & $0.708$  \\
      &      & $50$ & $\mathbf{0.745}$ & $0.701$ & $0.738$ & $0.734$ & $0.710$ & $0.696$  \\
      &      & $75$ & $\mathbf{0.740}$ & $0.692$ & $0.717$ & $0.719$ & $0.702$ & $0.678$ \\
\cmidrule{2-9}
      & $20$ & $30$ & $\mathbf{0.772}$ & $0.725$ & $0.754$ & $0.760$ & $0.736$ & $0.719$ \\
      &      & $50$ & $\mathbf{0.764}$ & $0.715$ & $0.740$ & $0.741$ & $0.728$ & $0.709$ \\
      &      & $75$ & $\mathbf{0.752}$ & $0.708$ & $0.728$ & $0.722$ & $0.719$ & $0.702$  \\
\midrule
$250$ & $10$ & $30$ & $\mathbf{0.779}$ & $0.744$ & $0.768$ & $0.760$ & $0.739$ & $0.732$ \\
      &      & $50$ & $\mathbf{0.774}$ & $0.734$ & $0.754$ & $0.747$ & $0.731$ & $0.726$ \\
      &      & $75$ & $\mathbf{0.763}$ & $0.720$ & $0.741$ & $0.732$ & $0.723$ & $0.715$  \\
\cmidrule{2-9}
      & $20$ & $30$ & $0.785$ & $0.748$ &  $0.769$ & $\mathbf{0.787}$ & $0.773$ & $0.744$ \\
      &      & $50$ & $\mathbf{0.781}$ & $0.736$ & $0.757$ & $0.761$ & $0.755$ & $0.738$  \\
      &      & $75$ & $\mathbf{0.769}$ & $0.732$ & $0.750$ & $0.747$ & $0.734$ & $0.722$ \\
\bottomrule
\end{tabular}
\caption{\small The results of classification accuracy over $50$ repeated experiments in Setting 1.}
\label{class_1}
\end{table}

\begin{table}[H]
\centering
\scriptsize
\begin{tabular}{@{}ccccccccc@{}}
\toprule
\textbf{Sample Size} & \textbf{Important Nodes} & \textbf{Nodes} & \textbf{MaGNet} & \textbf{PNGAT} & \textbf{GTN} & \textbf{GPS} & \textbf{MSGCN} & \textbf{APPNP} \\
\midrule
$100$ & $10$ & $30$ & $\textbf{0.753}$ & $0.718$ & $0.740$ & $0.728$ & $0.697$ & $0.746$  \\
 &  & $50$ & $\textbf{0.742}$ & $0.710$ & $0.729$ & $0.718$ & $0.684$ & $0.728$  \\
 &  & $75$ & $\textbf{0.736}$ & $0.697$ & $0.715$ & $0.710$ & $0.672$ & $0.709$ \\
\cmidrule{2-9} & $20$ & $30$ & $\textbf{0.768}$ & $0.740$ & $0.749$ & $0.762$ & $0.708$ & $0.728$ \\
 &  & $50$ & $0.755$ & $0.729$ & $0.744$ & $\textbf{0.760}$ & $0.701$ & $0.711$ \\
 &  & $75$ & $\textbf{0.748}$ & $0.710$ & $0.717$ & $0.719$ & $0.693$ & $0.704$  \\
\midrule $250$ & $10$ & $30$ & $0.776$ & $0.735$ & $\textbf{0.778}$ & $0.766$ & $0.728$ & $0.735$ \\
 &  & $50$ & $\textbf{0.771}$ & $0.733$ & $0.759$ & $0.758$ & $0.720$ & $0.727$ \\
 &  & $75$ & $\textbf{0.765}$ & $0.731$ & $0.740$ & $0.747$ & $0.711$ & $0.724$  \\
\cmidrule{2-9}  & $20$ & $30$ & $\textbf{0.786}$ & $0.754$ &  $0.763$ & $0.770$ & $0.767$ & $0.761$ \\
 &  & $50$ & $\textbf{0.779}$ & $0.750$ & $0.756$ & $0.759$ & $0.752$ & $0.753$  \\
 &  & $75$ & $\textbf{0.774}$ & $0.741$ & $0.752$ & $0.748$ & $0.734$ & $0.739$ \\
\bottomrule
\end{tabular}
\caption{\small The results of classification accuracy over $50$ repeated experiments in Setting 2.}
 \label{class_2}
\end{table}

As shown in Tables \ref{class_1} and \ref{class_2}, the MaGNet estimation model provides the best classification results among all the competing methods in general. This superior performance is consistent across varying sample sizes, node quantities, and important node sizes, indicating MaGNet's robust performance in graph classification tasks. This is mainly due to MaGNet's ability to effectively integrate both local and global information. The advantage of our model in solving the memoryless and over-smoothing issues results in effective and powerful representations for graph-structured data.

\subsection{Evaluation of the Interpretation Model}\label{sec:simu_explan}

In this section, we introduce different types of interpretation tasks and the corresponding results of the interpretation model. In particular, we consider three types of model interpretation tasks: node-wise, edge-wise, and feature-wise reasoning. We note that each of them is aligned with the functionalities of the proposed MaGNet interpretation model. In the following interpretation tasks, we consider a correlated and temporal data-generating process as in the simulation setting 2 in order to mimic the scenario with time-varying features in neural activity experiments.

In the node-wise interpretation tasks, the ultimate aim is to retain important nodes after node-wise reasoning. This is particularly crucial in practice for achieving a parsimonious model that simultaneously maintains interpretability and performance. In the edge-wise interpretation tasks, we initially define the notions of important and redundant edges (REs). An important edge refers to an edge connecting two important nodes. In contrast, all other edges not satisfying this condition are considered non-important edges. In the task, we seek to minimize the redundant edges in the trained MaGNet estimation model. To evaluate the interpretation model performance, we define two metrics: the absolute metric (AM) and the relative metric (RM), as follows: 
\begin{equation*}
\text{AM} := \frac{\char"0023\text{ of existing RE after reasoning}}{\char"0023\text{ of all possible RE}},
\end{equation*}
and
\begin{equation*}
\text{RM} := \frac{\char"0023\text{ of existing RE before reasoning} -  \char"0023\text{ of existing RE after reasoning}}{\char"0023\text{ of existing RE before reasoning }}.
\end{equation*} 
In the feature-wise interpretation tasks, we aim to detect influential features for graph nodes. In this task, we choose a submatrix with dimension $|V_0|\times p_0$ from the feature matrix $X$ for some $p_0 < p$. Then, we use this submatrix to form a classification rule as discussed above.

In the following, we present the results of the three types of interpretation task experiments. We compare our method to a state-of-the-art approach, Integrated Gradients (IntGradients, \cite{sundararajan2017axiomatic}), which is a gradient-based model interpretation approach designed for deep neural networks. For the implementation of IntGradients, we adapt the module of IntGradients to graph neural network settings and apply it to our trained MaGNet estimation model. Note that for the feature-wise interpretation tasks, we only evaluate the model performance of the MaGNet interpretation method, because IntGradients is not able to perform such type of task.

\begin{table}[!t]
\captionsetup{width=.7\linewidth}
\centering
\scriptsize
\begin{tabular}{@{}ccccc@{}}
\toprule
\textbf{Sample Size} & \textbf{Important Nodes} & \textbf{All Nodes} & \textbf{MaGNet} & \textbf{IntGradients} \\
\midrule
$100$ & $10$ & $30$ & $\textbf{0.763}$ & $0.708$ \\
      &     & $50$ & $\textbf{0.745}$   & $0.684$\\
\cmidrule{2-5} 
      & $20$ & $30$ & $\textbf{0.817}$  & $0.754$ \\
      &     & $50$ & $\textbf{0.795}$  & $0.752$\\
\midrule
$250$ & $10$ & $30$ & $\textbf{0.775}$ & $0.723$ \\
      &     & $50$ & $\textbf{0.758}$   & $0.687$\\
\cmidrule{2-5} 
      & $20$ & $30$ & $\textbf{0.826}$ & $0.767$ \\
      &     & $50$ & $\textbf{0.814}$ & $0.759$\\
\bottomrule
\end{tabular}
\caption{\small The node-wise interpretation performance over $50$ repeated experiments.}
\label{node_wise1}
\end{table}

\begin{table}[!t]
\captionsetup{width=.7\linewidth}
\centering
\scriptsize
\begin{tabular}{@{}ccccccc@{}}
\toprule
\textbf{Metric} & \textbf{Important Nodes} & \textbf{All Nodes} & \textbf{MaGNet}  & \textbf{IntGradients} \\ \midrule
\multirow{2}{*}{RM} & 10 & 30 & \textbf{0.806} & 0.768 \\
                    &    & 50 & \textbf{0.791} & 0.754 \\ \cmidrule(l){2-5} 
                    & 20 & 30 & \textbf{0.827} & 0.786 \\
                    &    & 50 & \textbf{0.812} & 0.771 \\  \midrule 
                    \multirow{2}{*}{AM} & 10 & 30 & \textbf{0.090} & 0.132 \\
                    &    & 50 & \textbf{0.128} & 0.161 \\ \cmidrule(l){2-5} 
                    & 20 & 30 & \textbf{0.056} & 0.101 \\
                    &    & 50 & \textbf{0.084} & 0.131 \\ \bottomrule
\end{tabular}
\caption{\small The edge-interpretation performance over two metrics with $50$ repeated experiments, where higher RM is better, and lower AM is better.}
\label{edge_wise1}
\end{table}

As shown in Table \ref{node_wise1}, the recovery rate of important nodes using the MaGNet interpretation model consistently outperforms the competing method across various settings. This is mainly due to the strength of our method in terms of leveraging the information gain to directly assess the reduction of uncertainty for the node subgraph. This technique incorporates statistical uncertainty as a measurement criterion instead of applying a fully deterministic gradient-based method. Another advantage of our method is the reparametrization strategy for continuously approximating discrete variables. This makes the proposed interpretation framework more computationally stable than the competing methods. Furthermore, the MaGNet interpretation model is particularly designed for the graph neural network and inherits the properties of the trained MaGNet estimation model. As a result, it is able to leverage local and global information to do the interpretation.
\begin{figure}[b!]
\centering     
  \captionsetup{width=.85\linewidth}
\subfigure[The dimension of temporal feature $p=25$.]{\includegraphics[width=68mm]{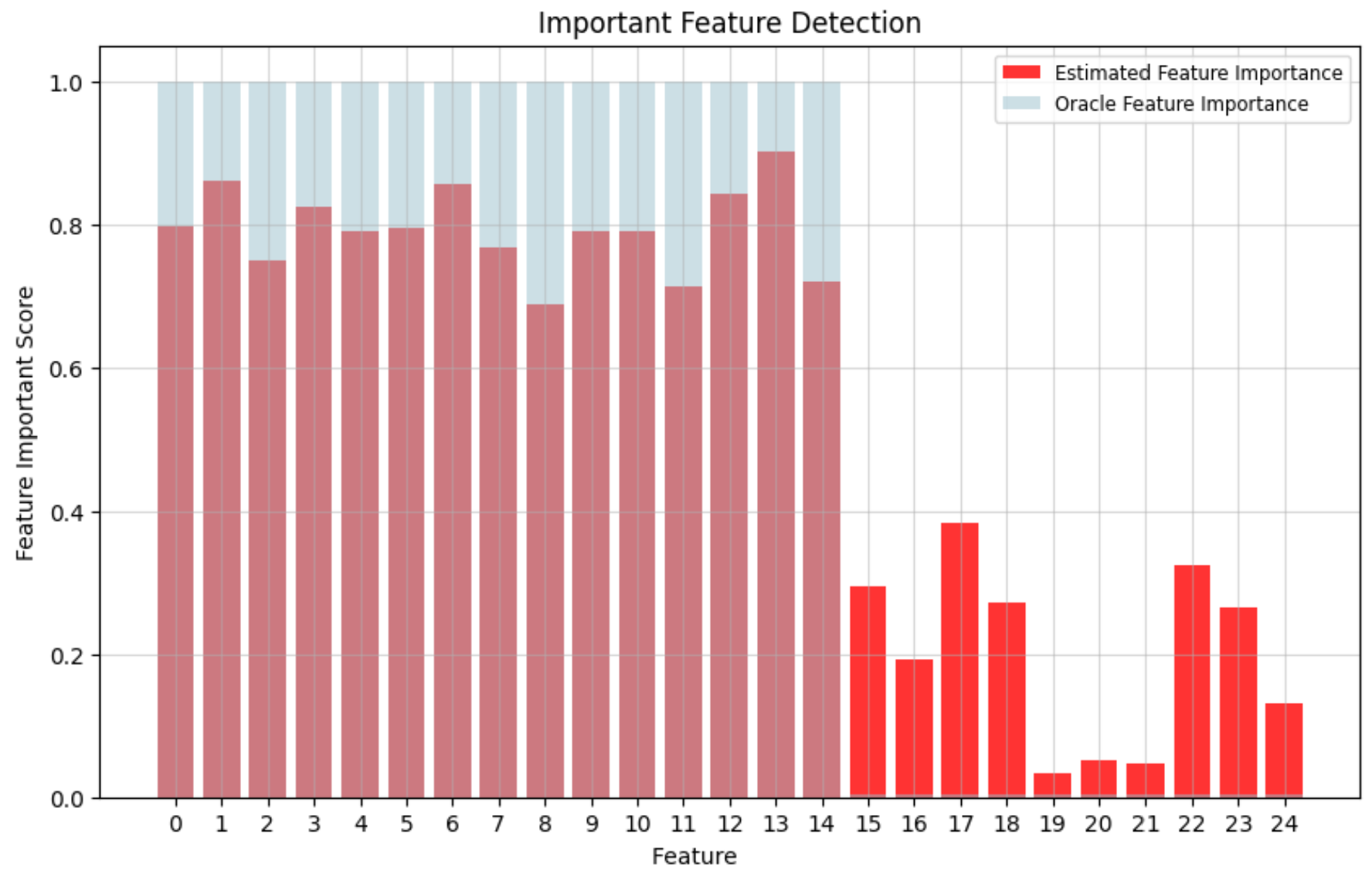}}
\subfigure[The dimension of temporal feature $p=50$.]{\includegraphics[width=68mm]{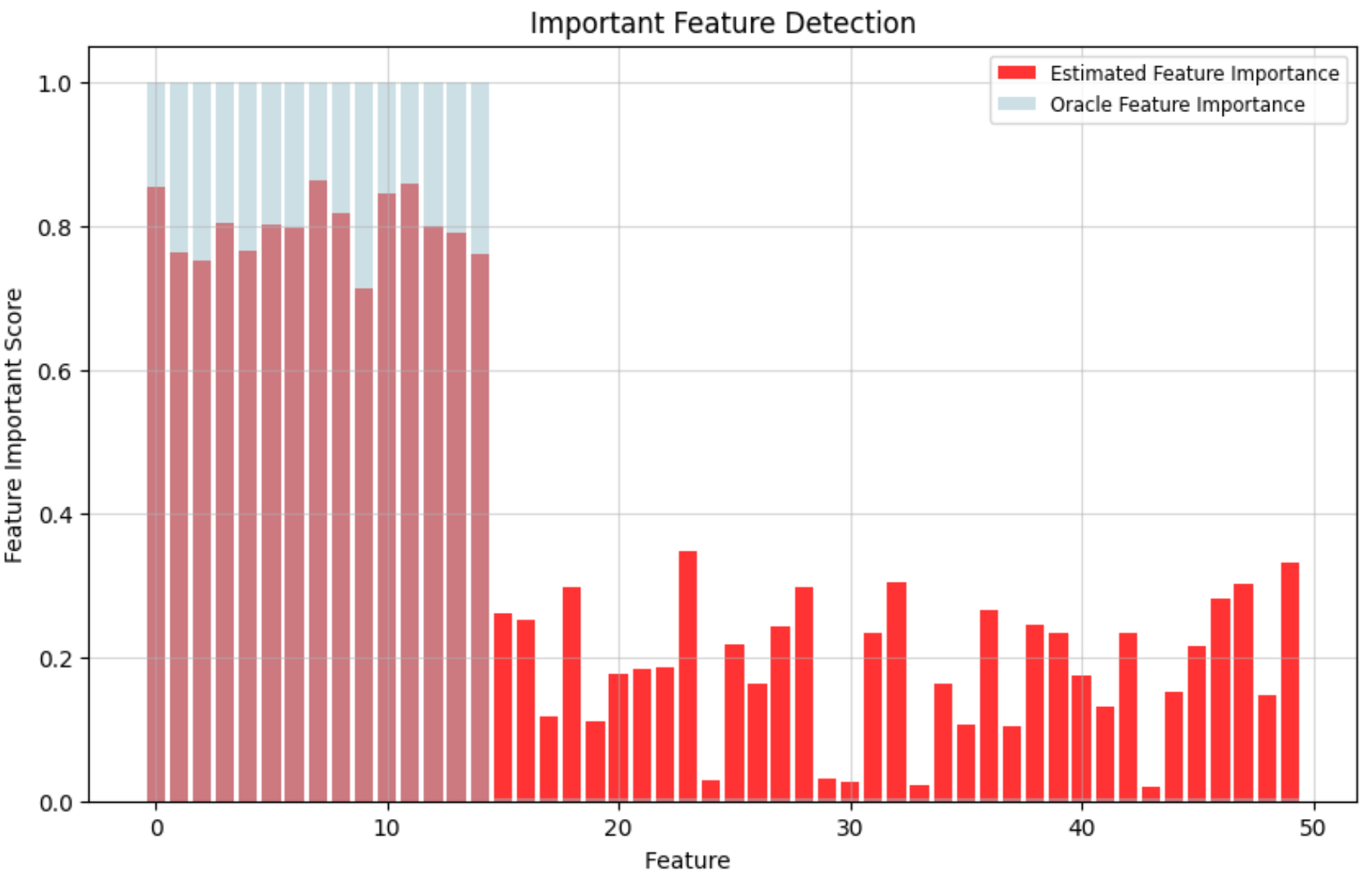}}
\caption{The feature-interpretation performance over $50$ repeated experiments}
    \label{fig:feat1_main}
    \end{figure}
The results of the edge-wise interpretation tasks are summarized in Table \ref{edge_wise1}.   
It shows that our proposed method has a high edge reduction rate, showcasing its effectiveness in pruning the non-important edges. Importantly, the method exhibits the ability to retain significant edges, suggesting an inherent capability in discriminating between important and non-important edges and, thus, preserving the influential subgraph structure. This performance is consistently validated across different settings, demonstrating the proposed method's robustness and adaptability to varying graph sizes. Further, these results demonstrate that our interpretation model can lead to a more parsimonious and interpretable results in practice. The results of the feature-wise interpretation task are reported in Figure \ref{fig:feat1_main}. The interpretation model tends to assign high scores to the top influential feature. It indicates that our model achieves great performance in temporal feature reasoning.

\section{Application to Local Field Potential Activity Data from the Rat Brain}
\label{real_sec}

In this section, we apply the proposed method to neural activity data recorded from an array of electrodes implanted inside the brain. The brain region of interest is the hippocampus, a region near the middle of the rat brain known to be important for the temporal organization of our memories and behaviors. Although it is well established that the hippocampus plays a key role in this function across mammals, the underlying neuronal mechanisms remain unclear. To shed light on these underlying mechanisms, we previously recorded neural activity in the hippocampus of rats performing a complex sequence memory task \citep{allen2016nonspatial}  (as such high-precision data are currently not available in humans). Using that dataset, our objective here is to apply the proposed method to identify key functional relationships in the local field potential (LFP) activity simultaneously recorded across electrodes during task performance, as this information could provide novel insights into potential functional relationships within that region.

\begin{figure}[h]
\centering	
\captionsetup{width=.8\linewidth}
\scalebox{0.21}[0.21]{\includegraphics{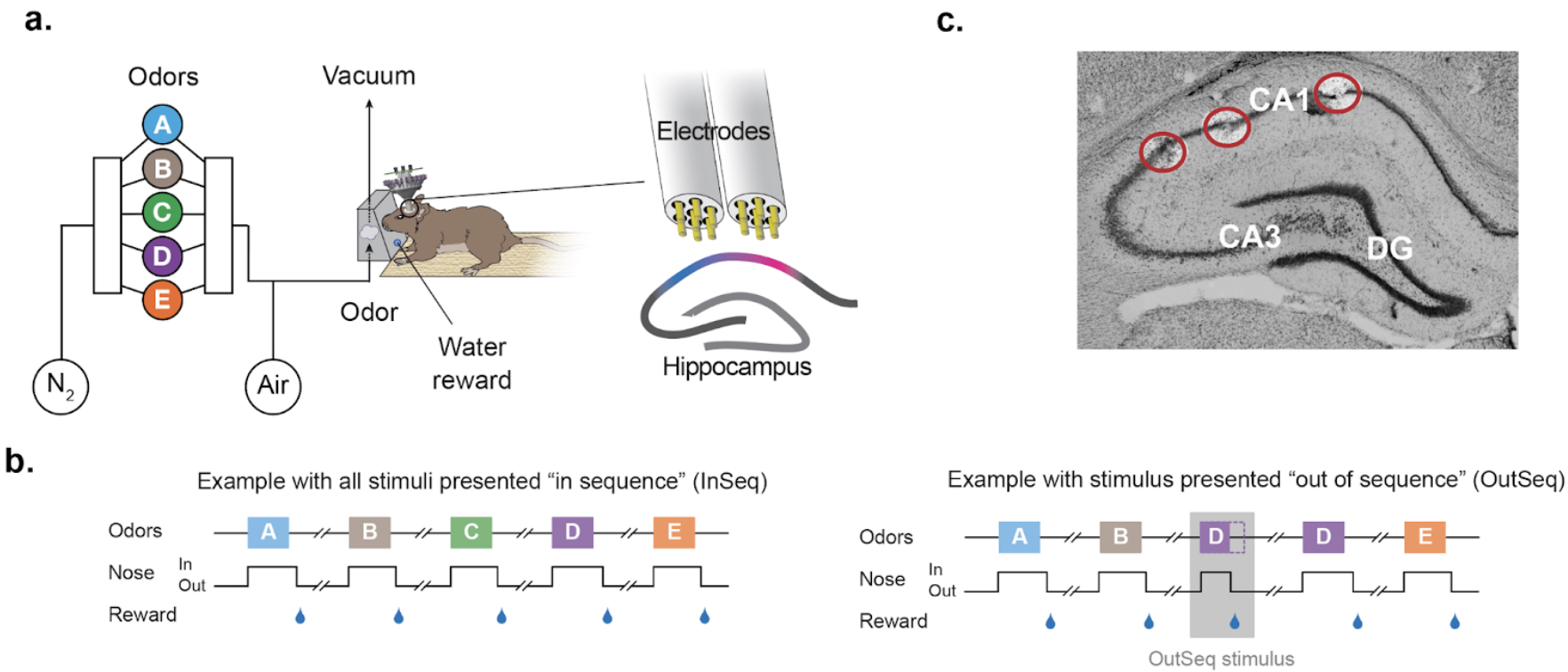}}
	\caption{
 \textbf{(a.)} The task involves repeated presentations of sequences of odors and requires rats to determine whether each odor was presented “in sequence” (InSeq; e.g., AB\underline{C}…) or “out of sequence” (OutSeq; e.g., AB\underline{D}…). Using an automated delivery system (left), all odors were presented in the same odor port (median interval between odors $\sim$5 s). Recordings were performed from electrodes organized into two bundles (right), which spanned much of the proximo-distal axis of dorsal CA1. \textbf{(b.)} In each session, the same sequence was presented multiple times, with approximately half the presentations including all InSeq trials (left) and the other half including one OutSeq trial (right). Each odor presentation was initiated by a nosepoke and rats were required to correctly identify each odor as either InSeq (by holding their nosepoke response until a tone signaled the end of the odor at 1.2 s) or OutSeq (by withdrawing their nose before the signal; <1.2 s) to receive a water reward. Incorrect responses resulted in the termination of the sequence. \textbf{(c.)} Location of three electrode tips (red circles). The leftmost and rightmost electrodes approximate the extent of the CA1 transverse axis recorded in each animal.
 }
 \label{fig:Behavior}
\end{figure}

The LFP neural activity data were collected from the CA1 region of the hippocampus while rats performed an odor sequence memory task (Figure  \ref{fig:Behavior}). In this task, rats received repeated presentations of odor sequences (e.g., ABCDE) at a single odor port and were required to identify each item as either “in sequence” (InSeq; e.g., AB\underline{C}…) or “out of sequence” (OutSeq; e.g., AB\underline{D}…). Importantly, the recordings were performed from surgically implanted electrodes (tetrodes), organized into two bundles, which spanned much of the proximo-distal axis of dorsal CA1. This experimental design thus provides a unique opportunity to directly examine the anatomical distribution of information processing along that axis. 

In recent work, \cite{shahbaba2022hippocampal} showed that information about trial content, such as the identity of the odor presented and whether it was presented in or out of sequence, could be accurately decoded from the ensemble spiking activity. However, that study did not determine whether task-relevant information was also contained in the local field potential activity. A fundamentally different data type from the discrete neural spiking activity, the LFP's continuous signal is more challenging to decode. To our knowledge, there are only two reports that exclusively use LFP to successfully decode spatial information in the hippocampus, which requires high-density recordings \citep{taxidis2015local, agarwal2014spatially}, and none showing decoding of nonspatial information from hippocampal LFP alone. To address this gap in knowledge, here we examined whether the content of odor trials can be decoded from hippocampal LFP activity and, if so, whether the dynamics vary over space (electrodes) and time.

\begin{figure}[H]
\centering	
\captionsetup{width=.7\linewidth}
\scalebox{0.41}[0.37]{\includegraphics{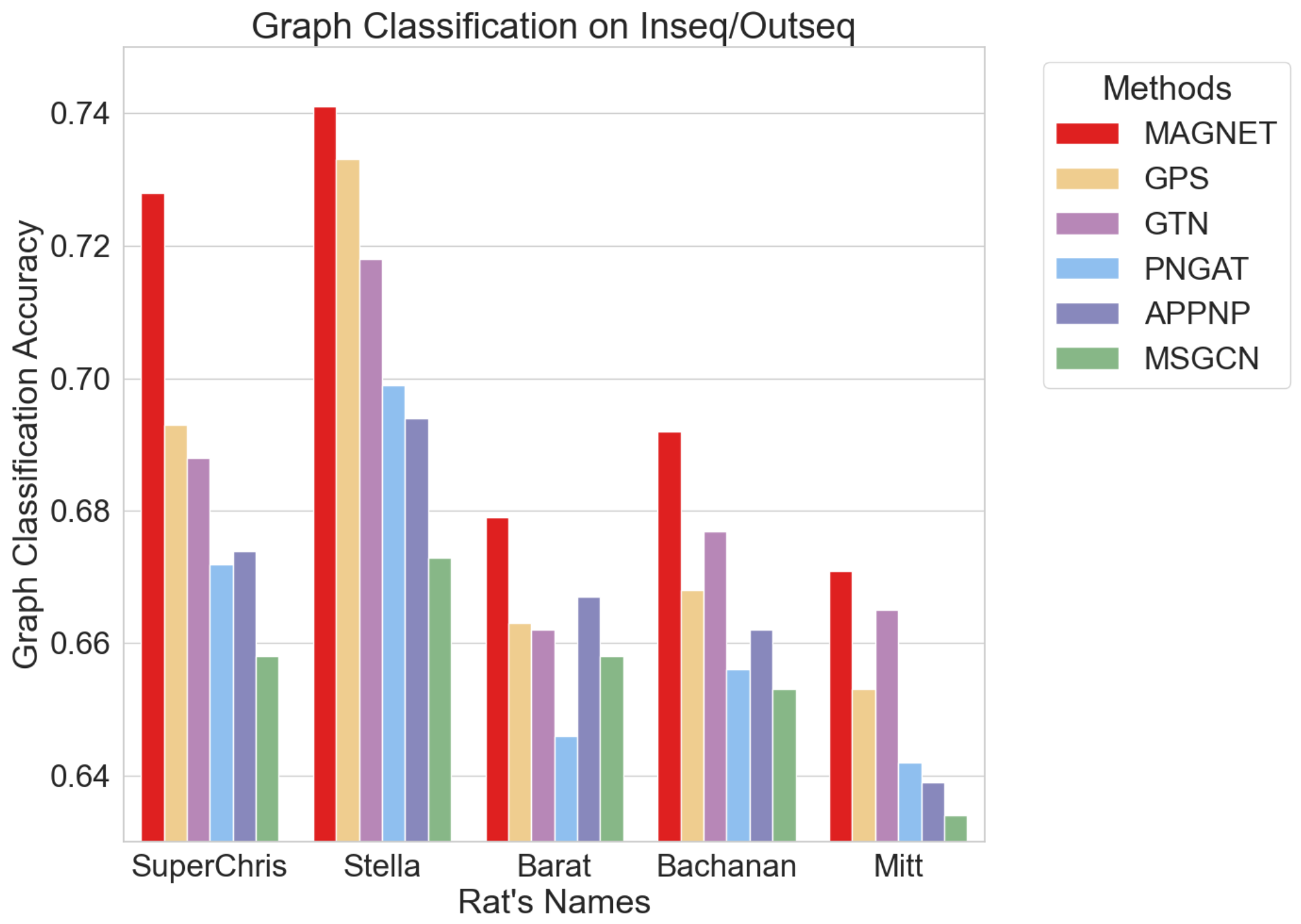}}
	\caption{Barplot of estimation accuracy for the MaGNet estimation model and alternative competing approaches on decoding the two main trial types.}\label{fig:feat1plot}
\end{figure}

For this analysis, we have focused on decoding the two main trial types (InSeq and OutSeq) using LFP activity from the 0-500 ms period (0 = odor onset), a time period in which there are no overt differences in the behavior of the animals between InSeq and OutSeq trials. We considered each rat's data an independent dataset and performed the classification evaluation task separately. For each rat's data, we randomly selected about $70$ graph instances as the training set and the other $30$ graph samples as the testing set. Figure \ref{fig:feat1plot} shows that the MaGNet estimation model achieves the best performance for all the rats. This is because our method's integration of both low-order and high-order information effectively utilizes more information into the latent representation. In addition, due to the proposed actor-critic structure, our method is less likely to suffer the \textit{over-smoothing} and \textit{memoryless} issue.

\begin{figure}[!t]
	\centering	
 \captionsetup{width=.75\linewidth}
 \scalebox{.76}[.76]{\includegraphics{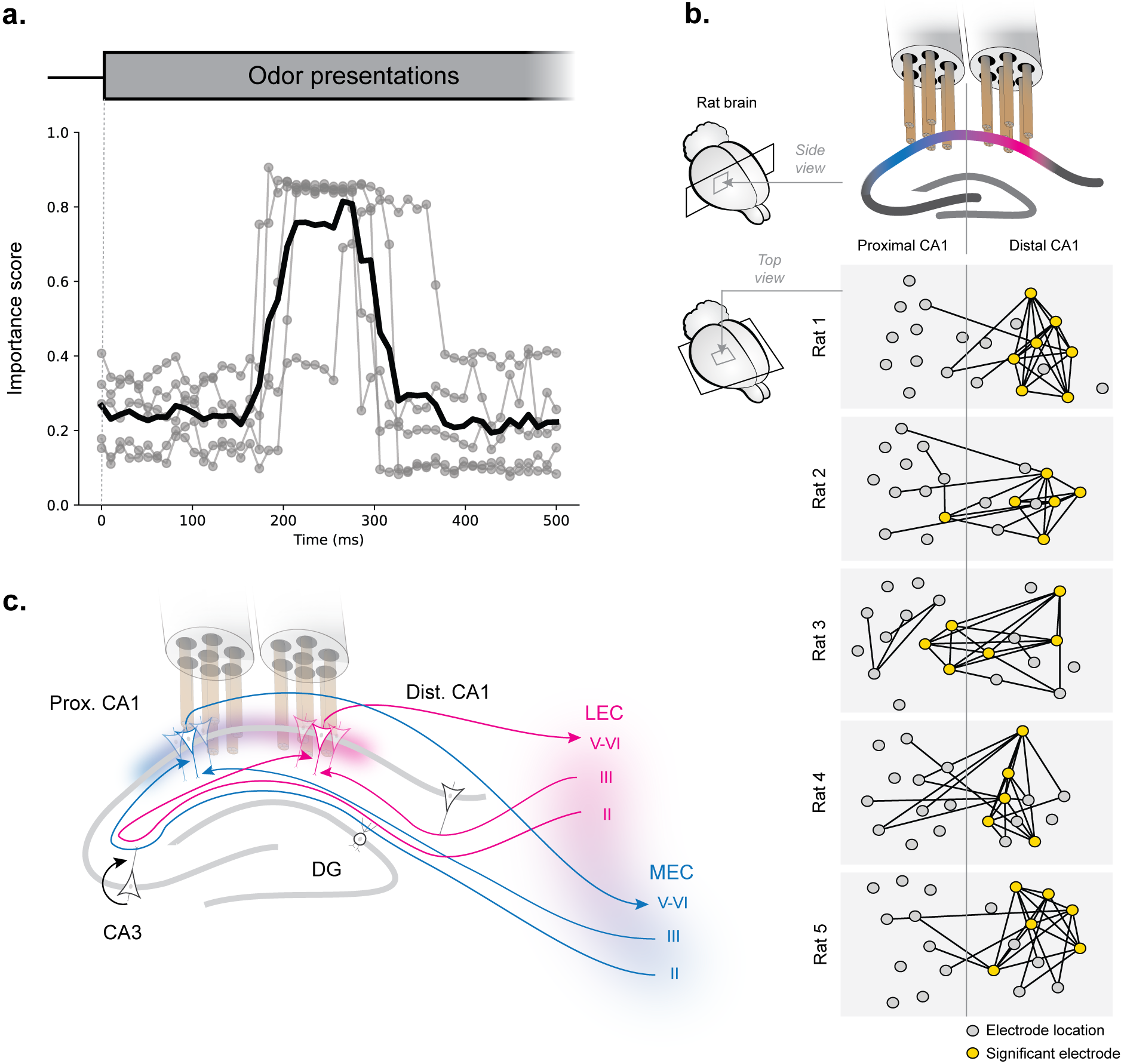}}
 \caption{
 \textbf{(a.)} Significant decoding of InSeq and OutSeq trials based on LFP activity during the first 500ms of odor trials. Scores peak during the 185-320 ms period, prior to the behavioral response. Grey traces indicate individual subject decodings, the black line indicates the mean across subjects.  
 \textbf{(b.)} Informative electrode nodes in the distal region of CA1. Schematic showing side view of electrode bundles implanted across the CA1 proximal-distal axis (Top). Schematic showing a top view of the anatomical distribution of electrodes across subjects based on electrode tract reconstruction (bottom). Yellow indicates significant nodes (electrodes).
 \textbf{(c.)} The clustering of informative nodes in distal CA1 is consistent with known anatomical differences in input connections. Odor information enters the hippocampus primarily through the LEC, which more strongly projects to the distal segment of CA1. In contrast, the MEC more strongly projects to proximal CA1. Approximate location of the implanted electrode bundles is shown.
}
\label{fig:EphysResults}
\end{figure}

We have also investigated the temporal dynamics of this decoding during trial periods by applying our MaGNet interpretation model. Specifically, we examined the most informative time bins \citep{allen2020hippocampus} for the InSeq/OutSeq classification in the first 500 ms of trials in Figure \ref{fig:EphysResults}a. We found that the most informative time bins occurred between $\sim$180 ms and $\sim$320 ms after the rats poked into the port. This timeline is consistent with reports of hippocampal neurons responding to odor information in as little as 100ms \citep{allen2020hippocampus} and with the expected timeline of InSeq/OutSeq identification within trials. With the neuropsychology's experimental knowledge, this implies that the MaGNet interpretation model is able to successfully identify the influential neural dynamics.

In addition, we have found that most informative electrodes clustered in the distal region of CA1 in Figure  \ref{fig:EphysResults}b. In fact, across all 5 rats, the majority ($86.7\%$) of significant electrodes were in distal CA1, and more than half of all electrodes in distal CA1 reached significance. This distribution of significant nodes suggests distal CA1 plays a more important role in representing InSeq/OutSeq information than proximal CA1, a pattern consistent with known differences in their anatomical connections (Figure \ref{fig:EphysResults}c).
Odor information enters the hippocampus primarily through the LEC (lateral entorhinal cortex), which more strongly projects to the distal segment of CA1. In contrast, the MEC (medial entorhinal cortex) more strongly projects to proximal CA1. However, the observation that some significant nodes also extended into proximal CA1 suggests that functional interactions among the two segments of CA1 are critical for task performance. In summary, we apply the MaGNet framework to LFP activity data recorded from the hippocampus of rats as they performed a challenging non-spatial sequence memory task. According to the analysis, we can decode the trial type (whether the odor was presented in or out of sequence) as well as identify the most informative trial periods and electrodes. Therefore, not only did the model provide the first direct evidence of decoding non-spatial trial content from hippocampal LFP activity alone, it also provided a high degree of specificity about how this information was distributed over space and time. This neuroscience result is consistent with a growing literature on the influence of anatomical gradients on information processing within brain regions \citep{Witter2017, Knierim2014}, specifically with evidence that inputs carrying non-spatial information more strongly project to distal CA1 than proximal CA1 \citep{Agster2009}.

\section{Discussion}

In this paper, we have proposed a novel graph neural network framework, MaGNet, which is able to effectively integrate both low-order and high-order information to allow powerful latent representation. Furthermore, MaGNet includes a practically useful interpretation component, which offers a tractable framework for identifying influential subgraphs, as well as important nodes, edges, and node features. In addition, we have also established rigorous theoretical foundations to assess the efficacy, statistical complexity, and generalizability of the MaGNet estimation model. These theoretical results ensure that the proposed estimation model is reliable and effective, contributing to the practical utility of MaGNet in various applications, especially in neuroscience.

One of the potential directions for exploration is to extend the current framework to accommodate different types of tasks. For example, rather than solely focusing on graph classification, the framework can be extended for node classifications, link prediction, and beyond. Furthermore, conducting rigorous theoretical investigations into the interpretation model -- such as studying how the introduced approximations and relaxations affect selection errors -- is both crucial and promising. Additionally, future research could explore studying the expressivity of the proposed method using the Weisfeiler-Lehman graph isomorphism test. Another potential research direction is to investigate generalization error in non-regular settings to provide a more comprehensive theoretical understanding of the current model. Exploring these directions will help expand the utility and effectiveness of our framework for a wide range of applications.

Moreover, there is a pressing need to expand the current framework to support dynamic settings. While modeling time-varying changes and dynamic systems holds central importance in numerous real-world applications, the current MaGNet framework (along with the majority of GNN models) is primarily tailored for static graph data. These models are capable of incorporating structural information into the learning process, but they fall short in capturing the evolution of dynamic graphs. Typically, dynamics in a graph refer to node attribute modifications or edge-structure changes, including the additions and deletions of nodes or edges. 
As a possible expansion of the existing MaGNet framework, we will explore the incorporation of node and edge activation functions to signify and capture the presence of the nodes and edges within each timestamp. This will enable subsequent utilization of attention mechanisms such as self-attention and neighborhood attention, which have shown efficacy in foundational models \citep{bommasani2021opportunities}, in order to account for historical time-evolved information from preceding timestamps.

\bibliographystyle{asa}
\bibliography{gnn_neuro.bib}

\newcommand{\noop}[1]{}
\begin{thebibliography}{51}
\newcommand{\enquote}[1]{``#1''}
\expandafter\ifx\csname natexlab\endcsname\relax\def\natexlab#1{#1}\fi

\bibitem[{Abu-El-Haija et~al.(2019)Abu-El-Haija, Perozzi, Kapoor, Alipourfard, Lerman, Harutyunyan, Ver~Steeg, and Galstyan}]{abu2019mixhop}
Abu-El-Haija, S., Perozzi, B., Kapoor, A., Alipourfard, N., Lerman, K., Harutyunyan, H., Ver~Steeg, G., and Galstyan, A. (2019), \enquote{Mixhop: Higher-order graph convolutional architectures via sparsified neighborhood mixing,} in \textit{International Conference on Machine Learning}, PMLR, pp. 21--29.

\bibitem[{Agarwal et~al.(2014)Agarwal, Stevenson, Ber{\'e}nyi, Mizuseki, Buzs{\'a}ki, and Sommer}]{agarwal2014spatially}
Agarwal, G., Stevenson, I.~H., Ber{\'e}nyi, A., Mizuseki, K., Buzs{\'a}ki, G., and Sommer, F.~T. (2014), \enquote{Spatially distributed local fields in the hippocampus encode rat position,} \textit{Science}, 344, 626--630.

\bibitem[{Agster and Burwell(2009)}]{Agster2009}
Agster, K.~L. and Burwell, R.~D. (2009), \enquote{{Cortical efferents of the perirhinal, postrhinal, and entorhinal cortices of the rat.}} \textit{Hippocampus}, 19, 1159--86.

\bibitem[{Allen et~al.(2020)Allen, Lesyshyn, O’Dell, Allen, and Fortin}]{allen2020hippocampus}
Allen, L.~M., Lesyshyn, R.~A., O’Dell, S.~J., Allen, T.~A., and Fortin, N.~J. (2020), \enquote{The hippocampus, prefrontal cortex, and perirhinal cortex are critical to incidental order memory,} \textit{Behavioural Brain Research}, 379, 112215.

\bibitem[{Allen et~al.(2016)Allen, Salz, McKenzie, and Fortin}]{allen2016nonspatial}
Allen, T.~A., Salz, D.~M., McKenzie, S., and Fortin, N.~J. (2016), \enquote{Nonspatial sequence coding in CA1 neurons,} \textit{Journal of Neuroscience}, 36, 1547--1563.

\bibitem[{Bodnar et~al.(2022)Bodnar, Di~Giovanni, Chamberlain, Lio, and Bronstein}]{bodnar2022neural}
Bodnar, C., Di~Giovanni, F., Chamberlain, B., Lio, P., and Bronstein, M. (2022), \enquote{Neural sheaf diffusion: A topological perspective on heterophily and oversmoothing in gnns,} \textit{Advances in Neural Information Processing Systems}, 35, 18527--18541.

\bibitem[{Bollob{\'a}s(1998)}]{bollobas1998modern}
Bollob{\'a}s, B. (1998), \textit{Modern Graph Theory}, vol. 184, Springer Science \& Business Media.

\bibitem[{Bommasani et~al.(2021)Bommasani, Hudson, Adeli, Altman, Arora, von Arx, Bernstein, Bohg, Bosselut, Brunskill, et~al.}]{bommasani2021opportunities}
Bommasani, R., Hudson, D.~A., Adeli, E., Altman, R., Arora, S., von Arx, S., Bernstein, M.~S., Bohg, J., Bosselut, A., Brunskill, E., et~al. (2021), \enquote{On the opportunities and risks of foundation models,} \textit{arXiv preprint arXiv:2108.07258}.

\bibitem[{Bottou(2010)}]{bottou2010large}
Bottou, L. (2010), \enquote{Large-scale machine learning with stochastic gradient descent,} in \textit{Proceedings of COMPSTAT'2010: 19th International Conference on Computational StatisticsParis France, August 22-27, 2010 Keynote, Invited and Contributed Papers}, Springer, pp. 177--186.

\bibitem[{Dabkowski and Gal(2017)}]{dabkowski2017real}
Dabkowski, P. and Gal, Y. (2017), \enquote{Real time image saliency for black box classifiers,} \textit{Advances in Neural Information Processing Systems}, 30.

\bibitem[{Devlin et~al.(2018)Devlin, Chang, Lee, and Toutanova}]{devlin2018bert}
Devlin, J., Chang, M.-W., Lee, K., and Toutanova, K. (2018), \enquote{Bert: Pre-training of deep bidirectional transformers for language understanding,} \textit{arXiv preprint arXiv:1810.04805}.

\bibitem[{Esser et~al.(2021)Esser, Chennuru~Vankadara, and Ghoshdastidar}]{esser2021learning}
Esser, P., Chennuru~Vankadara, L., and Ghoshdastidar, D. (2021), \enquote{Learning theory can (sometimes) explain generalisation in graph neural networks,} \textit{Advances in Neural Information Processing Systems}, 34, 27043--27056.

\bibitem[{Garg et~al.(2020)Garg, Jegelka, and Jaakkola}]{garg2020generalization}
Garg, V., Jegelka, S., and Jaakkola, T. (2020), \enquote{Generalization and representational limits of graph neural networks,} in \textit{International Conference on Machine Learning}, PMLR, pp. 3419--3430.

\bibitem[{Gasteiger et~al.(2018)Gasteiger, Bojchevski, and G{\"u}nnemann}]{gasteiger2018predict}
Gasteiger, J., Bojchevski, A., and G{\"u}nnemann, S. (2018), \enquote{Predict then Propagate: Graph Neural Networks meet Personalized PageRank,} in \textit{International Conference on Learning Representations}.

\bibitem[{Graham et~al.(2019)Graham, Wang, and Ravanbakhsh}]{graham2019equivariant}
Graham, D., Wang, J., and Ravanbakhsh, S. (2019), \enquote{Equivariant entity-relationship networks,} \textit{arXiv preprint arXiv:1903.09033}.

\bibitem[{Hamilton(2020)}]{hamilton2020graph}
Hamilton, W.~L. (2020), \enquote{Graph representation learning,} \textit{Synthesis Lectures on Artifical Intelligence and Machine Learning}, 14, 1--159.

\bibitem[{Ivanov and Prokhorenkova(2021)}]{ivanov2021boost}
Ivanov, S. and Prokhorenkova, L. (2021), \enquote{Boost then Convolve: Gradient Boosting Meets Graph Neural Networks,} in \textit{International Conference on Learning Representations}.

\bibitem[{Kipf et~al.(2018)Kipf, Fetaya, Wang, Welling, and Zemel}]{kipf2018neural}
Kipf, T., Fetaya, E., Wang, K.-C., Welling, M., and Zemel, R. (2018), \enquote{Neural relational inference for interacting systems,} in \textit{International conference on machine learning}, PMLR, p. 2688.

\bibitem[{Kipf and Welling(2016)}]{kipf2016semi}
Kipf, T.~N. and Welling, M. (2016), \enquote{Semi-supervised classification with graph convolutional networks,} \textit{arXiv preprint arXiv:1609.02907}.

\bibitem[{Knierim et~al.(2014)Knierim, Neunuebel, and Deshmukh}]{Knierim2014}
Knierim, J.~J., Neunuebel, J.~P., and Deshmukh, S.~S. (2014), \enquote{{Functional correlates of the lateral and medial entorhinal cortex: objects, path integration and local–global reference frames},} \textit{Philosophical Transactions of the Royal Society B: Biological Sciences}, 369, 20130369.

\bibitem[{Larose and Larose(2014)}]{larose2014discovering}
Larose, D.~T. and Larose, C.~D. (2014), \textit{Discovering knowledge in data: an introduction to data mining}, vol.~4, John Wiley \& Sons.

\bibitem[{Li et~al.(2021)Li, M{\"u}ller, Ghanem, and Koltun}]{li2021training}
Li, G., M{\"u}ller, M., Ghanem, B., and Koltun, V. (2021), \enquote{Training graph neural networks with 1000 layers,} in \textit{International conference on machine learning}, PMLR, pp. 6437--6449.

\bibitem[{Li et~al.(2018)Li, Han, and Wu}]{li2018deeper}
Li, Q., Han, Z., and Wu, X.-M. (2018), \enquote{Deeper insights into graph convolutional networks for semi-supervised learning,} in \textit{The AAAI conference on artificial intelligence}, vol.~32.

\bibitem[{Liao et~al.(2020)Liao, Urtasun, and Zemel}]{liao2020pac}
Liao, R., Urtasun, R., and Zemel, R. (2020), \enquote{A pac-bayesian approach to generalization bounds for graph neural networks,} \textit{arXiv preprint arXiv:2012.07690}.

\bibitem[{Liu et~al.(2022)Liu, Hooi, Kawaguchi, and Xiao}]{liu2022mgnni}
Liu, J., Hooi, B., Kawaguchi, K., and Xiao, X. (2022), \enquote{MGNNI: Multiscale Graph Neural Networks with Implicit Layers,} \textit{Advances in Neural Information Processing Systems}.

\bibitem[{Luo et~al.(2020)Luo, Cheng, Xu, Yu, Zong, Chen, and Zhang}]{luo2020parameterized}
Luo, D., Cheng, W., Xu, D., Yu, W., Zong, B., Chen, H., and Zhang, X. (2020), \enquote{Parameterized explainer for graph neural network,} \textit{Advances in neural information processing systems}, 33, 19620--19631.

\bibitem[{Lv(2021)}]{lv2021generalization}
Lv, S. (2021), \enquote{Generalization bounds for graph convolutional neural networks via rademacher complexity,} \textit{arXiv preprint arXiv:2102.10234}.

\bibitem[{Maddison et~al.(2016)Maddison, Mnih, and Teh}]{maddison2016concrete}
Maddison, C.~J., Mnih, A., and Teh, Y.~W. (2016), \enquote{The concrete distribution: A continuous relaxation of discrete random variables,} \textit{arXiv preprint arXiv:1611.00712}.

\bibitem[{Oono and Suzuki(2020)}]{oono2020optimization}
Oono, K. and Suzuki, T. (2020), \enquote{Optimization and generalization analysis of transduction through gradient boosting and application to multi-scale graph neural networks,} \textit{Advances in Neural Information Processing Systems}, 33, 18917--18930.

\bibitem[{Paulus et~al.(2020)Paulus, Choi, Tarlow, Krause, and Maddison}]{paulus2020gradient}
Paulus, M., Choi, D., Tarlow, D., Krause, A., and Maddison, C.~J. (2020), \enquote{Gradient estimation with stochastic softmax tricks,} \textit{Advances in Neural Information Processing Systems}, 33, 5691--5704.

\bibitem[{Ramp{\'a}{\v{s}}ek et~al.(2022)Ramp{\'a}{\v{s}}ek, Galkin, Dwivedi, Luu, Wolf, and Beaini}]{rampavsek2022recipe}
Ramp{\'a}{\v{s}}ek, L., Galkin, M., Dwivedi, V.~P., Luu, A.~T., Wolf, G., and Beaini, D. (2022), \enquote{Recipe for a general, powerful, scalable graph transformer,} \textit{Advances in Neural Information Processing Systems}, 35, 14501--14515.

\bibitem[{Reitzner et~al.(2017)Reitzner, Schulte, and Th{\"a}le}]{reitzner2017limit}
Reitzner, M., Schulte, M., and Th{\"a}le, C. (2017), \enquote{Limit theory for the Gilbert graph,} \textit{Advances in Applied Mathematics}, 88, 26--61.

\bibitem[{Scarselli et~al.(2018)Scarselli, Tsoi, and Hagenbuchner}]{scarselli2018vapnik}
Scarselli, F., Tsoi, A.~C., and Hagenbuchner, M. (2018), \enquote{The vapnik--chervonenkis dimension of graph and recursive neural networks,} \textit{Neural Networks}, 108, 248--259.

\bibitem[{Shahbaba et~al.(2022)Shahbaba, Li, Agostinelli, Saraf, Cooper, Haghverdian, Elias, Baldi, and Fortin}]{shahbaba2022hippocampal}
Shahbaba, B., Li, L., Agostinelli, F., Saraf, M., Cooper, K.~W., Haghverdian, D., Elias, G.~A., Baldi, P., and Fortin, N.~J. (2022), \enquote{Hippocampal ensembles represent sequential relationships among an extended sequence of nonspatial events,} \textit{Nature communications}, 13, 787.

\bibitem[{Sun et~al.(2019)Sun, Zhu, and Lin}]{sun2019adagcn}
Sun, K., Zhu, Z., and Lin, Z. (2019), \enquote{Adagcn: Adaboosting graph convolutional networks into deep models,} \textit{arXiv preprint arXiv:1908.05081}.

\bibitem[{Sundararajan et~al.(2017)Sundararajan, Taly, and Yan}]{sundararajan2017axiomatic}
Sundararajan, M., Taly, A., and Yan, Q. (2017), \enquote{Axiomatic attribution for deep networks,} in \textit{International conference on machine learning}, PMLR, pp. 3319--3328.

\bibitem[{Taxidis et~al.(2015)Taxidis, Anastassiou, Diba, and Koch}]{taxidis2015local}
Taxidis, J., Anastassiou, C.~A., Diba, K., and Koch, C. (2015), \enquote{Local field potentials encode place cell ensemble activation during hippocampal sharp wave ripples,} \textit{Neuron}, 87, 590--604.

\bibitem[{Veli{\v{c}}kovi{\'c}(2023)}]{velivckovic2023everything}
Veli{\v{c}}kovi{\'c}, P. (2023), \enquote{Everything is connected: Graph neural networks,} \textit{Current Opinion in Structural Biology}, 79, 102538.

\bibitem[{Veli{\v{c}}kovi{\'c} et~al.(2017)Veli{\v{c}}kovi{\'c}, Cucurull, Casanova, Romero, Lio, and Bengio}]{velivckovic2017graph}
Veli{\v{c}}kovi{\'c}, P., Cucurull, G., Casanova, A., Romero, A., Lio, P., and Bengio, Y. (2017), \enquote{Graph attention networks,} \textit{arXiv preprint arXiv:1710.10903}.

\bibitem[{Witter et~al.(2017)Witter, Doan, Jacobsen, Nilssen, and Ohara}]{Witter2017}
Witter, M.~P., Doan, T.~P., Jacobsen, B., Nilssen, E.~S., and Ohara, S. (2017), \enquote{{Architecture of the Entorhinal Cortex A Review of Entorhinal Anatomy in Rodents with Some Comparative Notes},} \textit{Frontiers in Systems Neuroscience}, 11, 46.

\bibitem[{Wu et~al.(2019)Wu, Souza, Zhang, Fifty, Yu, and Weinberger}]{wu2019simplifying}
Wu, F., Souza, A., Zhang, T., Fifty, C., Yu, T., and Weinberger, K. (2019), \enquote{Simplifying graph convolutional networks,} in \textit{International conference on machine learning}, PMLR, pp. 6861--6871.

\bibitem[{Wu et~al.(2021)Wu, Wang, Feng, He, Chen, Lian, and Xie}]{wu2021self}
Wu, J., Wang, X., Feng, F., He, X., Chen, L., Lian, J., and Xie, X. (2021), \enquote{Self-supervised graph learning for recommendation,} in \textit{Proceedings of the 44th international ACM SIGIR conference on research and development in information retrieval}, pp. 726--735.

\bibitem[{Wu et~al.(2020)Wu, Pan, Chen, Long, Zhang, and Yu}]{wu2020comprehensive}
Wu, Z., Pan, S., Chen, F., Long, G., Zhang, C., and Yu, P.~S. (2020), \enquote{A comprehensive survey on graph neural networks,} \textit{IEEE Transactions on Neural Networks and Learning Systems}, 32, 4--24.

\bibitem[{Xu et~al.(2018{\natexlab{a}})Xu, Hu, Leskovec, and Jegelka}]{xu2018powerful}
Xu, K., Hu, W., Leskovec, J., and Jegelka, S. (2018{\natexlab{a}}), \enquote{How powerful are graph neural networks?} \textit{arXiv preprint arXiv:1810.00826}.

\bibitem[{Xu et~al.(2018{\natexlab{b}})Xu, Li, Tian, Sonobe, Kawarabayashi, and Jegelka}]{xu2018representation}
Xu, K., Li, C., Tian, Y., Sonobe, T., Kawarabayashi, K.-i., and Jegelka, S. (2018{\natexlab{b}}), \enquote{Representation learning on graphs with jumping knowledge networks,} in \textit{International conference on machine learning}, PMLR, pp. 5453--5462.

\bibitem[{Yang et~al.(2020)Yang, Wang, Yao, Liu, and Abdelzaher}]{yang2020revisiting}
Yang, C., Wang, R., Yao, S., Liu, S., and Abdelzaher, T. (2020), \enquote{Revisiting over-smoothing in deep GCNs,} \textit{arXiv preprint arXiv:2003.13663}.

\bibitem[{Ying et~al.(2019)Ying, Bourgeois, You, Zitnik, and Leskovec}]{ying2019gnnexplainer}
Ying, Z., Bourgeois, D., You, J., Zitnik, M., and Leskovec, J. (2019), \enquote{Gnnexplainer: Generating explanations for graph neural networks,} \textit{Advances in neural information processing systems}, 32.

\bibitem[{Yun et~al.(2019)Yun, Jeong, Kim, Kang, and Kim}]{yun2019graph}
Yun, S., Jeong, M., Kim, R., Kang, J., and Kim, H.~J. (2019), \enquote{Graph transformer networks,} \textit{Advances in neural information processing systems}, 32.

\bibitem[{Zhang et~al.(2022)Zhang, Sheng, Yin, Jiang, Xia, Gao, Yang, and Cui}]{zhang2022model}
Zhang, W., Sheng, Z., Yin, Z., Jiang, Y., Xia, Y., Gao, J., Yang, Z., and Cui, B. (2022), \enquote{Model degradation hinders deep graph neural networks,} in \textit{Proceedings of the 28th ACM SIGKDD Conference on Knowledge Discovery and Data Mining}, pp. 2493--2503.

\bibitem[{Zhang et~al.(2020)Zhang, Cui, and Zhu}]{zhang2020deep}
Zhang, Z., Cui, P., and Zhu, W. (2020), \enquote{Deep learning on graphs: A survey,} \textit{IEEE Transactions on Knowledge and Data Engineering}, 34, 249--270.

\bibitem[{Zhao and Akoglu(2019)}]{zhao2019pairnorm}
Zhao, L. and Akoglu, L. (2019), \enquote{Pairnorm: Tackling oversmoothing in gnns,} \textit{arXiv preprint arXiv:1909.12223}.

\end{thebibliography}

\end{document}